\documentclass[journal]{IEEEtran}

\usepackage[utf8]{inputenc}
\usepackage[T1]{fontenc}

\usepackage{amsmath,amssymb,amsfonts}
\usepackage{nicefrac}
\usepackage{textcomp}

\usepackage{graphicx}
\graphicspath{{media/}}
\usepackage{booktabs}
\usepackage{makecell}
\usepackage{tabularx}
\usepackage{multirow}
\usepackage{float}
\usepackage{stfloats}
\usepackage[caption=false,font=normalsize,labelfont=sf,textfont=sf]{subfig}
\usepackage{pgfplots}
\pgfplotsset{compat=1.18}

\usepackage{algorithmic}
\usepackage{algorithm}
\usepackage{verbatim}

\usepackage{microtype}
\usepackage{pdflscape}
\usepackage{lipsum}
\usepackage{fancyhdr}
\usepackage[table]{xcolor}
\definecolor{red}{RGB}{255,0,0}
\definecolor{blue}{RGB}{0,0,255}
\definecolor{green}{RGB}{0,128,0}

\usepackage{url}
\usepackage{xr-hyper}
\usepackage{hyperref}
\hypersetup{
    hidelinks
}
\usepackage{cleveref}
\usepackage{cite}
\usepackage[acronym]{glossaries}
\makeglossaries
\input{glossary.glo}

\title{XCalib: Depth-Guided Geometric Optimization for Dense Thermal--Visible Video Registration}

\author{
Aur\'elien~Godet,
Gabriel~Jobert,
Mauro~Dalla Mura
\thanks{Aur\'elien Godet and Mauro Dalla Mura are with the Univ. Grenoble Alpes, CNRS, Grenoble INP--UGA, GIPSA-lab, 38000 Grenoble, France. Mauro Dalla Mura is also with the Institut Universitaire de France (IUF), France. Gabriel Jobert is with Lynred, 364 Avenue de Valence, 38113 Veurey-Voroize, France.}%
\thanks{This work was supported by the Fondation Grenoble INP, DeepRed chair, under the patronage of Lynred.}%
}

\begin{document}
\maketitle

\begin{abstract}
Image registration is a vital preprocessing step in multimodal perception tasks, including image fusion, object detection, and semantic segmentation. In Advanced Driver-Assistance Systems (ADAS), spatial misalignment between visible (RGB) and infrared (IR) cameras—caused by non-coincident optical axes and field-of-view differences—introduces non-uniform parallax and visual ghosting. Classical keypoint-based methods are restricted to global homographies that fail under dynamic depth, while unconstrained dense flow algorithms lack structural regularization and suffer from temporal instability.

In this paper, we propose XCalib, an unsupervised dense thermal-visible registration framework that bridges this gap. Rather than serving as an absolute metric calibration tool, XCalib leverages virtual pinhole camera parameterization strictly as a geometric constraint space. By optimizing effective relative pose and intrinsics alongside predicted monocular metric depth, XCalib restricts the search space of spatial displacements to physically valid projection geometries.

Our key contributions are: (1) a novel registration paradigm that uses camera parameterization as an implicit regularizer for dense cross-modal warping; (2) Normalized Edges Correlation (NEC), a robust structural similarity metric tailored to cross-spectral alignment; and (3) extensive quantitative and qualitative evaluations across public ADAS datasets, demonstrating superior temporal stability and alignment accuracy over unconstrained dense flow baselines.

The code will be made publicly available at \url{https://github.com/Azorgz/XCalib.git}.
\end{abstract}

\section{Introduction}
\label{sec:introduction}

Multimodal imaging systems combining visible light (RGB) and thermal infrared (IR) sensors are critical for robust environment perception in Advanced Driver-Assistance Systems (ADAS) and autonomous driving~\cite{jobert_2024}. RGB cameras capture high-resolution color and rich textural details under favorable illumination, whereas thermal sensors detect long-wave infrared radiation, offering resilience against poor lighting, fog, and severe headlight glare. Effective integration of these modalities enhances safety-critical visual tasks such as multi-spectral object detection, tracking, semantic segmentation, and image fusion.

However, downstream multimodal networks heavily rely on strict pixel-level spatial alignment~\cite{obj_det_depth, track_survey, seg_ADSMatching, survey2023}. In practical ADAS setups, cameras are mounted on multi-sensor rigs with non-coincident optical axes and distinct fields of view (FoV). This spatial separation induces depth-dependent parallax: near-field objects experience large coordinate displacements between sensors, while far-field objects shift minimally. Assuming pre-aligned image pairs or applying uniform planar homographies introduces severe ghosting artifacts and degrades downstream perception performance.

\begin{figure}[ht]
    \centering
    \includegraphics[width=0.95\linewidth]{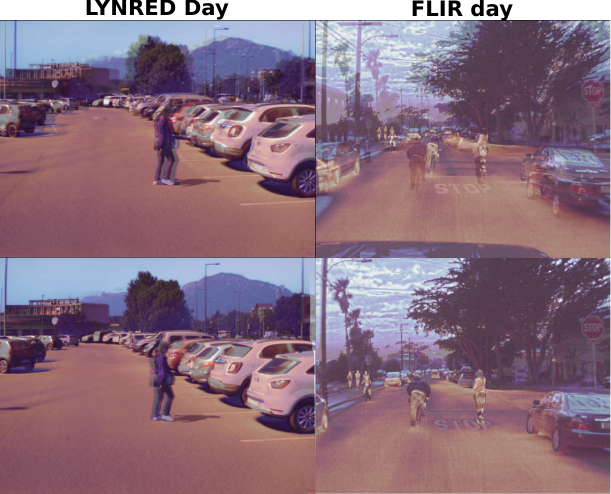}
    \caption{Samples from the Lynred Mobility Dataset (left column) and the FLIR ADAS v2 dataset (right column). 
    Top: superimposed visible and infrared images without registration. 
    Bottom: superimposed visible and warped infrared images using the dense mapping estimated by our method.}
    \label{fig:intro}
\end{figure}

Existing registration approaches generally fall between two paradigms: keypoint-based descriptor matching and unconstrained dense optical flow regression. Keypoint matching methods estimate global spatial transformations; while invariant to affine illumination changes, they cannot model depth-dependent parallax. Conversely, dense optical flow models regress pixel-wise displacement vectors directly. Although capable of handling local parallax, standard optical flow formulations lack structural geometric constraints. This unconstrained optimization often leads to localized structural deformations and severe temporal jitter across video sequences. While effective in some settings, both approaches face difficulties in cross-modal scenarios: feature descriptors often fail to generalize across modalities, and optical flow methods typically assume photometric consistency, which does not hold between RGB and IR images. We explore related work in more detail in Section~\ref{sec:Related Works}.

\begin{figure*}[ht]
    \centering
    \includegraphics[width=\linewidth]{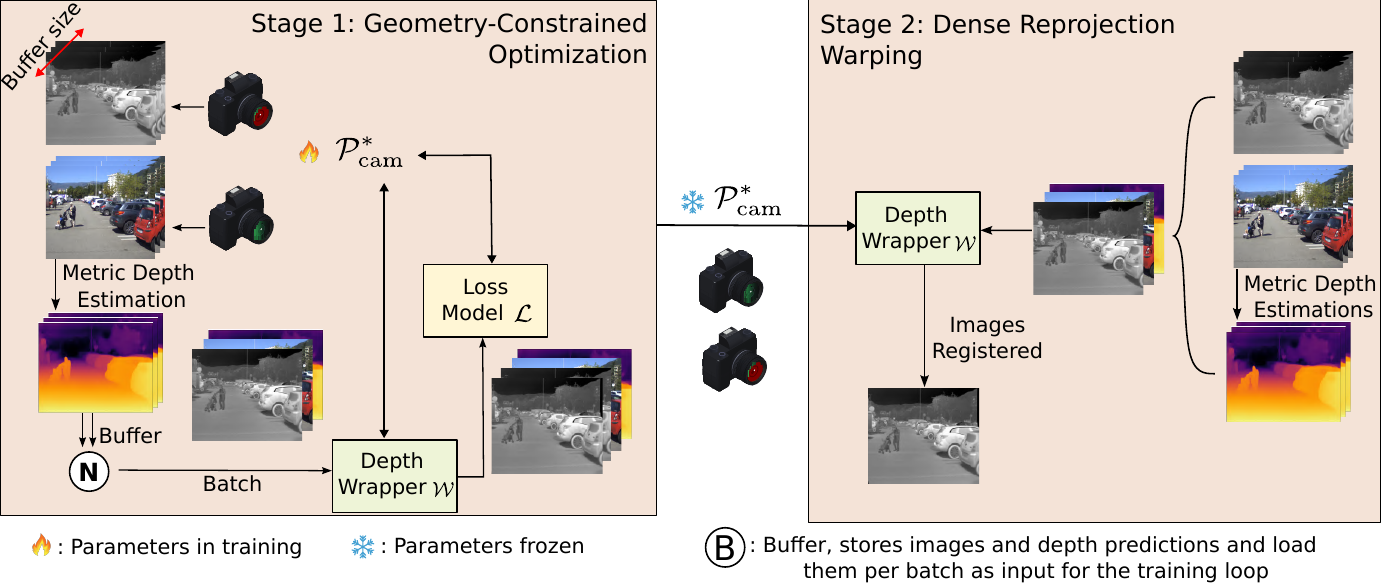}
    \caption{The proposed two-stage pipeline for thermal–visible image registration. The first stage involves unsupervised calibration of the camera rig to estimate intrinsic and extrinsic parameters using monocular depth estimation. The second stage uses these parameters, along with metric depth predictions, to compute dense pixel-wise mappings between thermal and visible images.}
    \label{fig:2stages}
\end{figure*}

To resolve this trade-off, we introduce \textbf{XCalib}, an unsupervised geometry-constrained registration framework. We emphasize that XCalib is not designed as a high-precision metric sensor calibration tool. Instead, it employs virtual pinhole camera intrinsics and extrinsics strictly as a geometric constraint space to parameterize dense reprojection flow fields. By modeling pixel shifts through relative projection equations and monocular metric depth, XCalib restricts the optimization search space to physically valid camera geometries without requiring accurate offline calibration targets or absolute ground-truth pose parameters. The core pipeline operates in two distinct optimization stages:

\begin{enumerate}
    \item \textbf{Geometry-Constrained Optimization:} We optimize an effective parameter set of intrinsic and extrinsic camera parameters using a combined objective comprising coarse cross-modal optical flow and a novel structural metric, \textbf{Normalized Edges Correlation (NEC)}. NEC evaluates edge magnitude and orientation correlation, providing strong optimization gradients across distinct spectral signatures.
    \item \textbf{Dense Reprojection Warping:} Using the optimized virtual parameter set and per-frame monocular metric depth maps, we compute dense non-rigid reprojection flow fields to warp source thermal images onto the reference visible coordinate lattice.
\end{enumerate}

By constraining dense displacement fields to mathematically valid camera projections, XCalib significantly mitigates unconstrained structural warping artifacts and enforces temporal stability across continuous video acquisitions. Experiments demonstrate that our approach produces smooth, temporally consistent registration flows on challenging ADAS datasets, matching or outperforming conventional keypoint-based and optical flow–based baselines. Although the method assumes a fixed multi-camera arrangement, the estimated parameter set remains valid for subsequent frame pairs acquired under the same setup.

\section{Related Works}
\label{sec:Related Works}

Accurate registration of thermal and visible images is a prerequisite for many multimodal perception and fusion tasks. However, the two modalities exhibit substantially different image formation processes and radiometric responses, making conventional intensity-based correspondence unreliable. In particular, corresponding structures may exhibit different contrast, missing texture, or even reversed edge polarity across modalities~\cite{reg_homo_keypointsPIIFD}. In addition, the spatial separation and different intrinsic parameters of the sensors introduce geometric discrepancies that cannot always be represented by a single global transformation. Existing thermal--visible registration methods can therefore be broadly divided into two main approaches: \textit{keypoint-based feature matching} and \textit{crossmodal dense flow estimation}.

\textbf{Keypoint-based Feature Matching.}
The first family identifies sparse correspondences between salient points or local image regions and uses them to estimate a spatial transformation. Classical descriptors such as SIFT~\cite{sift} were primarily designed for matching images with similar imaging characteristics and can therefore be unreliable across spectral modalities. In particular, gradient polarity and modality-dependent appearance variations can lead to ambiguous or incorrect correspondences. To address this issue, several descriptors specifically designed or adapted for multimodal registration have been proposed, including PIIFD~\cite{reg_homo_keypointsPIIFD}, DASC~\cite{reg_homo_keypoint_DASC} or RIFT~\cite{reg_homo_RIFT}. Other methods learn modality-invariant representations by extracting features from a shared latent space~\cite{reg_homo_features2022,reg_homo_features_Latentspace}. Recent approaches have also explored deep feature extractors followed by deformation of a regular image grid from sparse correspondences, enabling more flexible transformations than a purely rigid model~\cite{reg_keypoints_RA_MMIR}.

The main advantage of feature-based methods is their ability to establish correspondences while being relatively insensitive to modality-dependent intensity changes, scale, and orientation. However, the resulting correspondences remain sparse, and the transformation model used to propagate them to the full image generally imposes a global or smoothly varying deformation. For thermal--visible cameras with a non-negligible baseline, this assumption is restrictive: the displacement between corresponding pixels varies with scene depth, so a single homography or rigid transformation cannot simultaneously align objects located at substantially different distances. This limitation is particularly relevant for close-range ADAS scenes, where depth-dependent parallax can become significant.

\textbf{Crossmodal Flow Estimation.}
The second family directly estimates dense pixel correspondences between the two modalities. Optical flow has been extensively studied for conventional single-modality image sequences~\cite{Gibson1950,survey_flow}, and several works have adapted dense correspondence estimation to multimodal registration. NeMAR~\cite{reg_flow_NeMAR}, followed by UMF~\cite{reg_flow_UMF} and IMF~\cite{reg_flow_IMF}, introduces a modality-translation step to reduce the appearance discrepancy before estimating the deformation. VoxelMorph~\cite{VoxelMorph_2019} investigates unsupervised multimodal registration using a deep neural network and normalized cross-correlation (NCC) as a similarity objective. Other approaches estimate the crossmodal displacement more directly: SuperFusion~\cite{reg_flow_superfus} combines dense matching with semantic constraints, while CrossRAFT~\cite{reg_flow_crossRaft} adapts RAFT~\cite{raft_flow} to thermal--visible correspondence through complementary training and crossmodal data augmentation. These methods provide dense pixel-wise mappings and can therefore model local geometric variations that cannot be represented by a single global transformation.

Despite this flexibility, dense flow estimation also introduces a fundamental limitation for the considered problem. The predicted displacement field is generally optimized directly in image space, without explicitly encoding the persistent geometric relationship between the cameras. Consequently, the estimated mapping may contain local deformations that compensate for ambiguous or modality-specific image structures rather than reflecting the underlying sensor geometry. Moreover, when applied independently to consecutive frames, such methods have no explicit mechanism enforcing the temporal consistency expected from a fixed camera rig. Crossmodal dense matching is further challenged by large differences in scale, field of view, and appearance, which can make local feature correlation unreliable when the initial displacement is large.

\textbf{Geometry-constrained registration.}
The two families therefore expose a complementary trade-off. Feature-based approaches provide relatively robust correspondences but usually rely on a restrictive global transformation, whereas dense flow methods provide the required spatial flexibility at the cost of a largely unconstrained image-space deformation. An alternative is to constrain the dense mapping through the physical geometry of the sensor pair. In a fixed multi-camera rig, the relative camera parameters remain constant across acquisitions, while the image displacement varies primarily as a function of scene depth. A depth-aware camera model can therefore provide dense, depth-dependent correspondences while substantially reducing the degrees of freedom of the registration problem.

This principle has been explored in earlier thermal--visible video registration works through geometric transformations and depth or object-level constraints. For example, Torabi \textit{et al.}~\cite{torabi2012} estimate an affine transformation from trajectories for thermal--visible video registration, while LSS-based approaches incorporate disparity and depth-related cues to handle objects at different distances~\cite{torabi2013,torabi_LSS_BP}. These approaches demonstrate the benefit of exploiting geometric structure, but they either rely on global transformations or estimate correspondences/disparities directly rather than optimizing a persistent camera model. More recent deep-learning approaches have substantially improved dense crossmodal correspondence, but generally retain the image-space formulation described above.

Our approach combines the advantages of these two paradigms by parameterizing the dense displacement field through a shared camera configuration and monocular metric depth. Rather than estimating an unconstrained flow field independently for each frame pair, we optimize a compact set of intrinsic and extrinsic parameters and derive the dense mapping through the camera projection model. This formulation naturally introduces depth-dependent parallax while maintaining a consistent geometric configuration across acquisitions. The following section describes this formulation using a differentiable pinhole camera model and the corresponding depth-dependent reprojection.

\newcommand{\IIR}[1][red!70!black]{I_{\textcolor{#1}{\mathrm{IR}}}}
\newcommand{\tIIR}[1][red!70!black]{\widetilde{I}_{\textcolor{#1}{\mathrm{IR}}}}
\newcommand{\IVIS}[1][blue!70!black]{I_{\textcolor{#1}{\mathrm{VIS}}}}
\newcommand{\DVIS}[1][blue!70!black]{D_{\textcolor{#1}{\mathrm{VIS}}}}
\newcommand{\VIS}[1][]{{#1}_{\textcolor{blue!70!black}{\mathrm{VIS}}}}
\newcommand{\IR}[1][]{{#1}_{\textcolor{red!70!black}{\mathrm{IR}}}}
\newcommand{\CAM}[1][]{{#1}_{\textcolor{black}{\mathrm{CAM}}}}
\newcommand{\VIStoIR}[1]{\ensuremath{#1_{\textcolor{blue!70!black}{\mathrm{VIS}}}^{\textcolor{red!70!black}{\mathrm{IR}}}}}
\newcommand{\IRtoVIS}[1]{\ensuremath{#1_{\textcolor{red!70!black}{\mathrm{IR}}}^{\textcolor{blue!70!black}{\mathrm{VIS}}}}}
\newcommand{\RTVISIR}{\ensuremath{\left[\,R \mid t\,\right]_{\textcolor{blue!70!black}{\mathrm{VIS}}}^{\textcolor{red!70!black}{\mathrm{IR}}}}}
\newcommand{\RTWVIS}{\ensuremath{\left[\,R \mid t\,\right]_{\textcolor{black}{\mathrm{World}}}^{\textcolor{blue!70!black}{\mathrm{VIS}}}}}
\newcommand{\RTVISW}{\ensuremath{\left[\,R \mid t\,\right]_{\textcolor{blue!70!black}{\mathrm{VIS}}}^{\textcolor{black}{\mathrm{World}}}}}
\newcommand{\RTWIR}{\ensuremath{\left[\,R \mid t\,\right]_{\textcolor{black}{\mathrm{World}}}^{\textcolor{red!70!black}{\mathrm{IR}}}}}
\newcommand{\param}[1][]{\ensuremath{\widetilde{#1}}}
\newcommand{\RTWCAM}{\ensuremath{\left[\,R \mid t\,\right]_{\textcolor{black}{\mathrm{World}}}^{\textcolor{black}{\mathrm{CAM}}}}}
\section{Methodology}
\label{sec:methodology}

\subsection{Problem Formulation and Framework}
\label{sec:framework}

We consider a fixed multi-camera imaging system composed of a reference camera and a source camera. For each synchronized acquisition, the two cameras observe the same scene from different viewpoints and potentially through different sensing modalities. The objective is to estimate a dense spatial transformation mapping the pixels of the reference image onto their corresponding locations in the source image, thereby enabling the source image to be resampled on the reference image grid.

Let $\Omega_R \subset \mathbb{R}^2$ and $\Omega_S \subset \mathbb{R}^2$ denote the continuous image-coordinate domains of the reference and source cameras, respectively, with corresponding discrete pixel lattices $\Lambda_R$ and $\Lambda_S$. A pixel coordinate $\hat{\mathbf{x}} \in \Lambda_R$ represents a discrete sample location, whereas $\mathbf{x} \in \Omega_R$ denotes a continuous image coordinate.

The reference and source images are defined as
\begin{equation}
I_R : \Lambda_R \rightarrow \mathbb{R}^{d_R}, \qquad I_S : \Lambda_S \rightarrow \mathbb{R}^{d_S},
\end{equation}
where $d_R$ and $d_S$ denote their respective numbers of channels. In our visible--infrared configuration, the visible image acts as the reference image and the infrared image as the source image, though the formulation itself is modality-independent.

The proposed approach explicitly models the spatial transformation through the geometry of the camera system rather than estimating an independent displacement at every pixel. Let $\mathcal{P}_{\mathrm{cam}}$ denote the set of intrinsic and extrinsic camera parameters. Together with a depth map $D$ defined in the reference camera frame, these parameters induce a continuous mapping
\begin{equation}
\mathcal{T}_p : \Omega_R \rightarrow \Omega_S.
\label{eq:continuous_mapping}
\end{equation}

For a reference coordinate $\mathbf{x} \in \Omega_R$, the corresponding source coordinate is
\begin{equation}
\mathbf{x}_S = \mathcal{T}_p \left( \mathbf{x}; \mathcal{P}_{\mathrm{cam}}, D(\mathbf{x}) \right).
\end{equation}

Rather than explicitly optimizing a dense displacement field, the displacement is implicitly induced by the camera geometry and scene depth. For a discrete reference pixel $\hat{\mathbf{x}} \in \Lambda_R$, the induced displacement field is
\begin{equation}
\mathbf{u}(\hat{\mathbf{x}}) = \mathcal{T}_p \left( \hat{\mathbf{x}}; \mathcal{P}_{\mathrm{cam}}, D(\hat{\mathbf{x}}) \right) - \hat{\mathbf{x}}, \qquad \mathbf{u} : \Lambda_R \rightarrow \mathbb{R}^{2}.
\end{equation}

The dense transformation can therefore be written as
\begin{equation}
\mathcal{T}_p(\hat{\mathbf{x}}; \mathbf{u}) = \hat{\mathbf{x}} + \mathbf{u}(\hat{\mathbf{x}}), \qquad \mathbf{u} : \Lambda_R \rightarrow \mathbb{R}^{2}.
\label{eq:dense_mapping}
\end{equation}

Importantly, $\mathbf{u}$ is not represented by independent learnable variables; it is generated by a low-dimensional camera model and the depth associated with each reference pixel. Consequently, the resulting dense transformation is constrained to remain compatible with a physically realizable camera configuration. The following sections establish this geometric mapping and formulate its estimation as an unsupervised optimization problem.

\subsection{Geometric Formulation of Dense Registration}
\label{sec:geometric_model}

\subsubsection{Differentiable Geometric Mapping}
\label{sec:geometric_mapping}

We assume a pinhole camera model for both imaging systems. The geometric configuration is characterized by intrinsic matrices $\mathbf{K}_R$ and $\mathbf{K}_S$, and by the rigid transformation from the reference to the source camera coordinate system. The complete set of camera parameters is
\begin{equation}
\mathcal{P}_{\mathrm{cam}} = \left\{ \mathbf{K}_R, \mathbf{K}_S, \mathbf{R}, \mathbf{t} \right\},
\label{eq:camera_parameters}
\end{equation}
where $\mathbf{R} \in SO(3)$ is the relative rotation and $\mathbf{t} \in \mathbb{R}^3$ is the relative translation.

Let $\hat{\mathbf{x}} = [u, v]^{\mathsf T} \in \Lambda_R$ be a discrete reference pixel with homogeneous representation $\widetilde{\mathbf{x}} = [u, v, 1]^{\mathsf T}$. A depth map $D : \Lambda_R \rightarrow \mathbb{R}_{>0}$ provides the depth of each reference pixel in the reference frame. The depth map is treated as a fixed input to the proposed optimization, independently of whether it is obtained from a monocular metric depth estimator or an external measurement system.

Given $D(\hat{\mathbf{x}})$, the reference pixel is first unprojected into the 3D reference camera frame:
\begin{equation}
\mathbf{X}_R = D(\hat{\mathbf{x}}) \mathbf{K}_R^{-1} \widetilde{\mathbf{x}}.
\label{eq:unprojection}
\end{equation}

The 3D point is then transformed into the source camera coordinate system:
\begin{equation}
\mathbf{X}_S = \mathbf{R} \mathbf{X}_R + \mathbf{t}.
\label{eq:rigid_transform}
\end{equation}

Finally, perspective projection yields the corresponding continuous source-image coordinate:
\begin{equation}
\mathbf{x}_S=\pi\left(\mathbf{K}_S\mathbf{X}_S\right),
\label{eq:perspective_projection}
\end{equation}
where the perspective division operator $\pi\left([X,Y,Z]^{\mathsf T}\right)=[X/Z,Y/Z]^{\mathsf T}$ maps a 3D point in camera coordinates onto the continuous 2D image plane.

Combining Eqs.~\eqref{eq:unprojection}--\eqref{eq:perspective_projection} yields the dense geometric mapping: 
\begin{equation} 
\mathcal{T}_p \left( \hat{\mathbf{x}}; \mathcal{P}_{\mathrm{cam}}, D(\hat{\mathbf{x}}) \right) = \pi \left[ \mathbf{K}_S \left( D(\hat{\mathbf{x}}) \mathbf{R} \mathbf{K}_R^{-1} \widetilde{\mathbf{x}} + \mathbf{t} \right) \right]. 
\label{eq:3d_mapping} 
\end{equation}

Since perspective projection is invariant to non-zero uniform scaling, dividing the continuous mapping by $D(\hat{\mathbf{x}})$ yields the decomposed form:
\begin{equation} 
\mathcal{T}_p \left( \hat{\mathbf{x}}; \mathcal{P}_{\mathrm{cam}}, D(\hat{\mathbf{x}}) \right) = \pi_p \left( \underbrace{\vphantom{\frac{1}{D(\hat{\mathbf{x}})}} \mathbf{K}_S \mathbf{R} \mathbf{K}_R^{-1}}_{(1)} \widetilde{\mathbf{x}} + \underbrace{\frac{1}{D(\hat{\mathbf{x}})} \mathbf{K}_S}_{(2)} \mathbf{t} \right).
\label{eq:3d_wrapper} 
\end{equation}

Here, term~(1) represents the infinite homography matrix:
\begin{equation}
\mathbf{H}_{\infty} = \mathbf{K}_S \mathbf{R} \mathbf{K}_R^{-1},
\label{eq:infinite_homography}
\end{equation}
which is governed solely by the relative camera rotation and intrinsic parameters. Term~(2) introduces a parallax displacement scaling inversely with depth, $1/D(\hat{\mathbf{x}})$, thereby modeling the scene's 3D structure. Unlike a planar homography, this formulation accounts for spatially non-uniform displacements induced by camera translation and finite scene geometry as established in epipolar geometry literature~\cite{2D_7points_hartley_2003,2D_szeliski2022computer}.

The continuous source coordinates generally do not fall on integer lattice points of $\Lambda_S$. A differentiable resampling operator $\mathcal{B}(I_S, \mathbf{x}) : \Omega_S \rightarrow \mathbb{R}^{d_S}$ interpolates the discrete samples of $I_S$. The warped source image defined on $\Lambda_R$ is thus
\begin{equation}
I_S'(\hat{\mathbf{x}}) = \mathcal{B} \left( I_S, \mathcal{T}_p \left( \hat{\mathbf{x}}; \mathcal{P}_{\mathrm{cam}}, D(\hat{\mathbf{x}}) \right) \right), \qquad \hat{\mathbf{x}} \in \Lambda_R.
\label{eq:warping}
\end{equation}

We implement $\mathcal{B}$ using differentiable bilinear interpolation~\cite{registration_jaderberg_2016_spatialtransformernetworks}, establishing an end-to-end differentiable path from the registration loss to the camera parameters.

\subsubsection{Optimization Objective}
\label{sec:geometric_optimization}

Given a dataset of $N$ synchronized observations randomly sampled from the entire dataset to improve scene diversity, $\mathcal{D} = \{ (I_R^{(i)}, I_S^{(i)}, D^{(i)}) \}_{i=1}^{N}$, the camera parameters are estimated via
\begin{equation}
\mathcal{P}_{\mathrm{cam}}^{*} = \arg\min_{\mathcal{P}_{\mathrm{cam}}} \sum_{i=1}^{N} \mathcal{L} \left( I_R^{(i)}, I_S'^{(i)} \right)
\label{eq:dense_optimization}
\end{equation}
with
\begin{equation}
I_S'^{(i)} = \mathcal{W} \left( I_S^{(i)}, \mathcal{T}_p \left( \cdot; \mathcal{P}_{\mathrm{cam}}, D^{(i)} \right) \right),
\end{equation}
where $\mathcal{W}$ denotes the differentiable warping operation defined by Eq.~\eqref{eq:warping}.

The depth maps $D^{(i)}$ remain fixed throughout optimization. The method conditions camera configuration estimation on externally provided scene geometry rather than performing joint depth estimation and calibration.

\subsection{Registration Loss Model}
\label{sec:loss_model}

In the absence of ground-truth correspondences or baseline calibration, we formulate registration in an unsupervised manner using three complementary loss terms:
\begin{enumerate}
    \item a cross-modal optical-flow loss for coarse global alignment;
    \item an edge-based image similarity loss for fine structural alignment;
    \item a camera regularization loss constraining parameters to plausible configurations.
\end{enumerate}

\subsubsection{Cross-Modal Flow Loss}
\label{sec:flow_loss}

To estimate coarse displacement robustly across spectral modalities, we use the pretrained crossmodal flow estimator (Here CrossRAFT model~\cite{reg_flow_crossRaft}, but any suitable model can be used). Let $\mathbf{F} = \operatorname{CrossRAFT}(I_R, I_S') : \Lambda_R \rightarrow \mathbb{R}^{2}$ denote the residual flow field. The flow loss is defined as its mean Euclidean magnitude:
\begin{equation}
\mathcal{L}_{\mathrm{flow}} = \frac{1}{|\Lambda_R|} \sum_{\hat{\mathbf{x}} \in \Lambda_R} \|\mathbf{F}(\hat{\mathbf{x}})\|_2.
\label{eq:flow_loss}
\end{equation}

\subsubsection{Cross-Modal Edge Similarity}
\label{sec:image_loss}

Because pixel-wise intensity metrics (e.g., MSE, SSIM) fail across visible--infrared modalities due to radiometric differences, we evaluate structural similarity using Normalized Edge Correlation (NEC), which constitutes one contribution of this work. It's a differentiable, polarity-invariant edge similarity metric that combines gradient magnitude and orientation information.

For an image $I$, its gradient field $G(I) = [g(I), \theta(I)]^{\mathsf T}$ comprises magnitude and orientation components:
\begin{equation}
g(I) = \sqrt{(\partial_x I)^2 + (\partial_y I)^2}, \quad \theta(I) = \operatorname{atan2}\left(\partial_y I, \partial_x I\right).
\end{equation}

To suppress noise while limiting dominant edge artifacts, the magnitude is clamped via
\begin{equation}
g'(I) = \max \left( \min \left( g(I), 5\overline{g} \right), \frac{2\overline{g}}{r} \right),
\label{eq:gradient_clamping}
\end{equation}
where $\overline{g}$ denotes the mean gradient magnitude over $\Lambda_R$, and $r$ is the proportion of active gradient pixels.
Orientation alignment is measured using polarity-invariant absolute cosine similarity:
\begin{equation}
C(\hat{\mathbf{x}}) = \left| \cos \left( \theta(I_R)(\hat{\mathbf{x}}) - \theta(I_S')(\hat{\mathbf{x}}) \right) \right|.
\label{eq:gradient_orientation}
\end{equation}

The resulting Normalized Edge Correlation is
\begin{equation}
\mathrm{NEC} = \frac{\sum_{\hat{\mathbf{x}} \in \Lambda_R} g'(I_R)(\hat{\mathbf{x}}) g'(I_S')(\hat{\mathbf{x}}) C(\hat{\mathbf{x}})}{\sqrt{ \sum_{\hat{\mathbf{x}} \in \Lambda_R} g'(I_R)(\hat{\mathbf{x}})^2 \sum_{\hat{\mathbf{x}} \in \Lambda_R} g'(I_S')(\hat{\mathbf{x}})^2}}.
\label{eq:nec}
\end{equation}

Each pair's contribution is weighted by its structural information magnitude factor:
\begin{equation}
c_i = \sqrt{ \sum_{\hat{\mathbf{x}} \in \Lambda_R} g'(I_R^{(i)})(\hat{\mathbf{x}})^2 \sum_{\hat{\mathbf{x}} \in \Lambda_R} g'(I_S'^{(i)})(\hat{\mathbf{x}})^2}.
\label{eq:nec_coefficient}
\end{equation}

Over a batch of $N$ pairs, the image loss is
\begin{equation}
\mathcal{L}_{\mathrm{image}} = \frac{\sum_{i=1}^{N} c_i \left[ 1 - \mathrm{NEC} \left( I_R^{(i)}, I_S'^{(i)} \right) \right]}{\sum_{i=1}^{N} c_i}.
\label{eq:image_loss}
\end{equation}

\subsubsection{Camera Regularization}
\label{sec:regularization_loss}

To prevent parameter drift and coupling non-uniqueness~\cite{sun2026nodrift3rraymapguidedcouplingdriftrobust}, we penalize deviations from the initial state using the Huber loss:
\begin{equation}
\operatorname{HL}(a, b; \delta) = \begin{cases}
\frac{1}{2}(a - b)^2, & |a - b| \leq \delta, \\
\delta \left( |a - b| - \frac{1}{2}\delta \right), & |a - b| > \delta.
\end{cases}
\label{eq:huber}
\end{equation}

The regularization term on translation $\mathbf{t}$ and unit quaternion $\mathbf{Q}$ is
\begin{equation}
\mathcal{L}_{\mathrm{reg}} = \operatorname{HL}(\mathbf{t}, \mathbf{t}_0; 0.1) + \operatorname{HL}(\mathbf{Q}, \mathbf{Q}_0; 0.05).
\label{eq:camera_regularization}
\end{equation}

\subsubsection{Total Registration Objective}
\label{sec:total_loss}

The overall optimization objective is
\begin{equation}
\boxed{ \mathcal{L}_{\mathrm{total}} = \mathcal{L}_{\mathrm{image}} + \lambda \mathcal{L}_{\mathrm{flow}} + \gamma \mathcal{L}_{\mathrm{reg}} },
\label{eq:total_loss}
\end{equation}

\subsection{Differentiable Camera Parameterization}
\label{sec:camera_parameterization}

Direct optimization of raw physical parameters is ill-conditioned due to differing units and scales. We parameterize physical settings via unconstrained variables $\boldsymbol{\phi}$ through a differentiable mapping $\mathcal{P}_{\mathrm{cam}} = \Psi(\boldsymbol{\phi})$, yielding the execution chain:
\begin{equation}
\boldsymbol{\phi} \longrightarrow \mathcal{P}_{\mathrm{cam}} \longrightarrow \mathcal{T}_p \longrightarrow I_S' \longrightarrow \mathcal{L}_{\mathrm{total}}.
\label{eq:differentiable_chain}
\end{equation}

\subsubsection{Intrinsic Parameters}
\label{sec:intrinsics}

The intrinsic matrix $\mathbf{K}$ is parameterized as
\begin{equation}
\mathbf{K} = \begin{pmatrix} f_x & s & c_x \\ 0 & f_y & c_y \\ 0 & 0 & 1 \end{pmatrix},
\label{eq:intrinsic}
\end{equation}
where principal points are normalized to image dimensions $(w, h)$ via $c_x = \widetilde{c}_x w$, $c_y = \widetilde{c}_y h$ (initialized at $\widetilde{c}_{x_0} = \widetilde{c}_{y_0} = 0.5$), and skew is $s = \widetilde{s}$ ($\widetilde{s}_0 = 0$).

Focal lengths are parameterized via horizontal and vertical fields of view (initialized at a baseline $45^\circ$ HFoV, and scaled to the image aspect ratio):
\begin{equation}
    f_x = \frac{w}{2\tan \!\left( \frac{\widetilde{f}_x \pi}{8} \right)},\quad
    f_y = \frac{h}{2\tan \!\left[ \widetilde{f}_y \arctan \!\left( \frac{h}{w} \tan \!\left( \frac{\pi}{8} \right) \right) \right]}, 
    \label{eq:focal_lengths}
\end{equation}
where $\widetilde{f}_x$ and $\widetilde{f}_y$ are unconstrained optimization variables initialized at $1.0$.

\subsubsection{Extrinsic Parameters}
\label{sec:extrinsics}

Extrinsic pose is represented as $\mathbf{E} = [\mathbf{R} \mid \mathbf{t}]$. To enforce valid rotation constraints, rotation is parameterized via an unconstrained quaternion $\widetilde{\mathbf{Q}} = [\widetilde{q}_\theta, \widetilde{q}_x, \widetilde{q}_y, \widetilde{q}_z]^{\mathsf T}$, normalized as $\mathbf{Q} = \widetilde{\mathbf{Q}} / \|\widetilde{\mathbf{Q}}\|_2$ (initialized at $[1, 0, 0, 0]^{\mathsf T}$). Given $\mathbf{Q} = [q_\theta, q_x, q_y, q_z]^{\mathsf T}$, the rotation matrix is
\begin{equation}
\mathbf{R} = \begin{bmatrix}
1 - 2(q_y^2 + q_z^2) & 2(q_x q_y - q_\theta q_z) & 2(q_x q_z + q_\theta q_y) \\
2(q_x q_y + q_\theta q_z) & 1 - 2(q_x^2 + q_z^2) & 2(q_y q_z - q_\theta q_x) \\
2(q_x q_z - q_\theta q_y) & 2(q_y q_z + q_\theta q_x) & 1 - 2(q_x^2 + q_y^2)
\end{bmatrix}.
\label{eq:rotation_matrix}
\end{equation}

\subsubsection{Multi-Scale Translation Parameterization}
\label{sec:translation}

To account for scale disparities between parameters, translation is split into fine and coarse optimization components:
\begin{equation}
\mathbf{t} = \begin{bmatrix} \widetilde{t}^{[0]}_x \\ \widetilde{t}^{[0]}_y \\ \widetilde{t}^{[0]}_z \end{bmatrix} + \lambda_t \begin{bmatrix} \widetilde{t}^{[1]}_x \\ \widetilde{t}^{[1]}_y \\ \widetilde{t}^{[1]}_z \end{bmatrix},
\label{eq:translation}
\end{equation}
initialized at zero with scale factor $\lambda_t = 10$.




\subsection{Optimization Strategy}
\label{sec:training_strategy}

The camera parameters are not equally identifiable from the reprojection objective, and several of them can produce partially equivalent image-space effects. In particular, focal length and $z$-translation can both modify the apparent scale of the reprojection, while image-plane translation and the principal point can partially compensate for each other. Similar couplings arise between rotation and translation parameters. Optimizing all parameters simultaneously from the initial configuration can therefore lead to competing parameter updates, unstable trajectories, and convergence toward unfavorable local minima. This issue is further amplified by the uncertainty of the monocular depth estimates used to compute the reprojection.

To mitigate these effects, we adopt a two-stage optimization strategy with progressive parameter unfreezing. The first stage restricts the optimization to the dominant parameters, providing a stable initial geometric configuration while allowing the more strongly coupled or less influential parameters to remain fixed. The second stage then releases the remaining degrees of freedom for fine refinement around this configuration.

\begin{itemize}
\item \textbf{Stage 1 (Coarse Alignment):}
The dominant camera parameters are optimized using
$\mathcal{L}^{(1)} = \mathcal{L}_{\mathrm{image}} + \lambda \mathcal{L}_{\mathrm{flow}} + \gamma \mathcal{L}_{\mathrm{reg}}$,
with an initial learning rate $\eta_1 = 5 \times 10^{-3}$ and a decay rate of $0.98$ per step. Secondary and strongly coupled variables, namely $z$-translation, the principal point, roll, and skew, are frozen during this stage. This restriction reduces the effective dimensionality of the optimization and prevents these parameters from compensating prematurely for errors in the dominant geometric configuration. In particular, the $z$-translation has limited practical importance for the considered fixed camera rig compared with the lateral translation and rotational components.

\item \textbf{Stage 2 (Fine Refinement):}
At $75\%$ of the optimization schedule, all parameters are unfrozen and jointly refined. The regularization term is removed, yielding
$\mathcal{L}^{(2)} = \mathcal{L}_{\mathrm{image}} + \lambda \mathcal{L}_{\mathrm{flow}}$,
and the learning rate is reduced to $\eta_2 = 5 \times 10^{-4}$. At this point, the dominant parameters have already established a suitable geometric configuration, allowing the remaining variables to refine the alignment without inducing large compensating changes in the overall projection.
\end{itemize}

Optimization is performed with Adam for 15 epochs on a 32-frame buffer, using a batch size of 4. The resulting parameter set $\mathcal{P}_{\mathrm{cam}}^*$ defines a consistent virtual camera configuration and is directly reused for subsequent acquisitions from the same fixed camera rig. 
The hyperparameters of the loss are set with a grid search to $\lambda = 2.5$ and $\gamma = 1.0$.
\newcommand{\cmark}{\textcolor{green!50!black}{\checkmark}}
\newcommand{\xmark}{\textcolor{red!70!black}{\times}}
\newcommand{\first}[1]{\textbf{\textcolor{green}{#1}}}
\newcommand{\second}[1]{\underline{\textcolor{blue}{#1}}}
\newcommand{\third}[1]{\textit{\textcolor{cyan}{#1}}}

\section{Experiments}
\label{sec:experiments}
\noindent

In this section, we present five different experiments:
\begin{itemize}
    \item \textbf{(1)} A comparison of the registration capability of several common metrics used as loss functions against the proposed \acrshort{nec} loss.
    \item \textbf{(2)} An evaluation of the temporal consistency of the registration over video sequences.
    \item \textbf{(3)} A quantitative and qualitative evaluation of our method against selected state-of-the-art cross-modality registration algorithms.
    \item \textbf{(4)} A model analysis for the proposed multi-scale translation parametrization.
    \item \textbf{(5)} An ablation study to evaluate the impact of each component of our loss model (Presented in supplementary material).
    \item \textbf{(6)} A monocular depth noise and scale robustness analysis of the proposed framework (Presented in supplementary material). 
    
\end{itemize}
The two last experiments are detailed in the supplementary material. The first three experiments are presented in this section.
\vspace{5pt}
\noindent \textbf{State-of-The-Art algorithms:} To evaluate the efficiency of our method, we compare the alignment results with other \Acrfull{sota} algorithms. The selected methods are: 
\begin{itemize}
    \item \textbf{IMF}~\cite{reg_flow_IMF}, an improved version of UMF~\cite{reg_flow_UMF}, both of which rely on modality translation followed by the estimation of a deformation field for image registration.
    \item \textbf{SuperFusion}~\cite{reg_flow_superfus}, which directly estimates a deformation field by computing cross features correlation between the two modalities.
    \item \textbf{PGMR}~\cite{zheng_plug-and-play_2025}, a recent method inspired by SuperFusion with the addition of a modality adapation module and an result enhancer.
    \item \textbf{CrossRAFT}~\cite{reg_flow_crossRaft}, which estimates the optical flow directly between the two modalities. This is a clever adaptation of the RAFT~\cite{raft_flow} architecture for cross-modality flow.
    \item \textbf{RIFT}~\cite{reg_homo_RIFT}, a classical keypoint matching algorithm made for cross-modality keypoints matching. We do not expect this algorithm to perform as well as the other as the final wrap is a rigid homography, not fit for correcting parallax effects.
\end{itemize}

\begin{figure}[htbp]
    \begin{center}
        \includegraphics[width=0.95\linewidth]{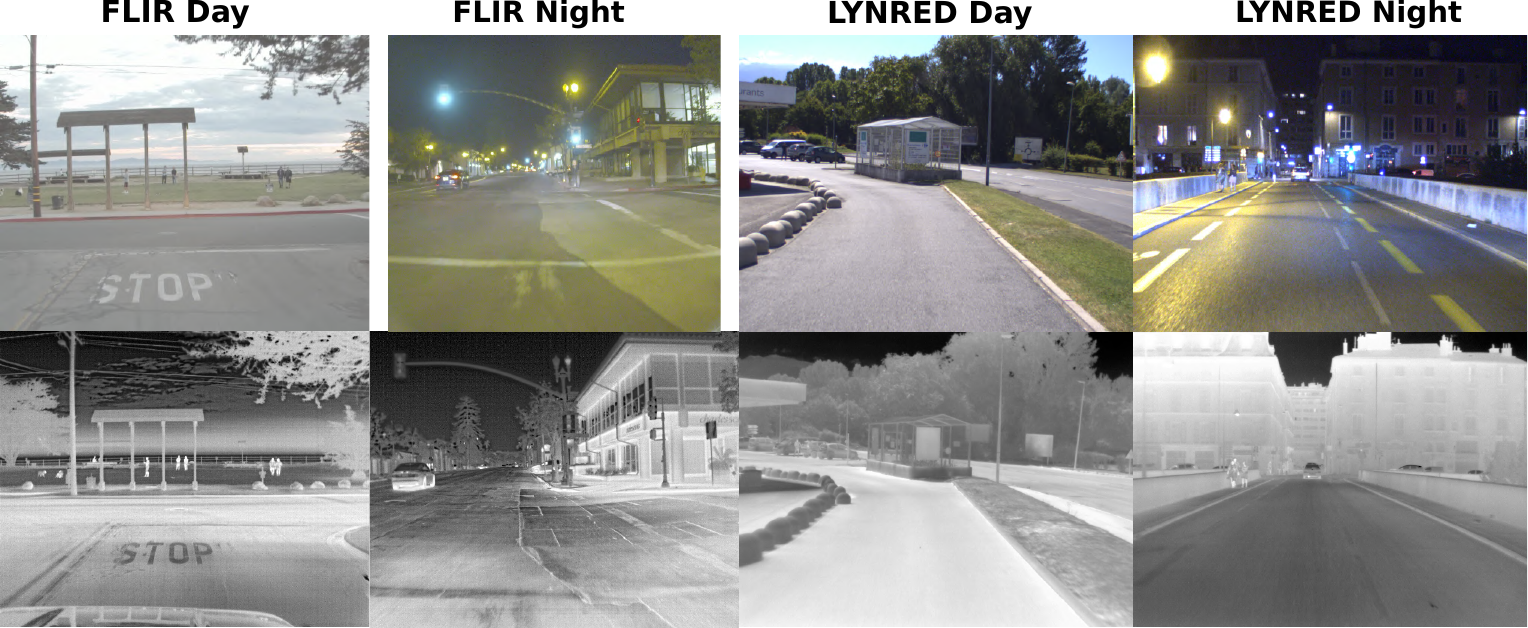}
    \end{center}
    \caption{Samples from the selected datasets. First row: the rgb images, Second row: the infrared images}
    \label{fig:mosaic_dataset}
\end{figure}

\noindent \textbf{Datasets:} The experiments are conducted on a selection of image sequences chosen based on their difficulty (see \cref{fig:mosaic_dataset}). Sample extracts from each dataset are presented in \cref{fig:mosaic_dataset}.

\begin{itemize}
    \item \textbf{FLIR day/night} – Respectively a day/night sequence of 672/1040 pairs of visible and thermal images extracted from the FLIR ADAS v2 dataset~\cite{flirDataset}.
    \item \textbf{Lynred day/night} – 2 Sequences of 1800 pairs of left visible and left thermal images from the Lynred mobility dataset~\cite{lynred_lynred_2025}.
    \item \textbf{RoadScene} – 221 aligned pairs of visible and thermal images ~\cite{xu2020aaai}. We use it only for the first experiment as it's already aligned.
\end{itemize}

\begin{table}[H]
    \caption{Datasets used for the experiments.
    *The color rates the difficulty of the alignment: \textcolor{green}{easy}, \textcolor{orange}{medium}, \textcolor{red}{high}.
    **Visibility: refers to the illumation or the weather conditions}
    \centering
    \begin{tabular}{|l|c|c|c|c|c|c|}
    \hline
    Dataset*                         & Vis**   & \makecell{Focal\\diff} & Baseline & \makecell{Resolution\\RGB} & \makecell{Nb\\frames} \\
    \hline\hline
    \textcolor{green}{Lynred day}    & high    & low                    & low      & 960x1280             & 1800 \\
    \textcolor{green}{Lynred night}  & med    & low                    & low      & 960x1280             & 1800 \\
    \textcolor{orange}{FLIR day}     & high    & med                    & low      & 1024x1280            & 672 \\
    \textcolor{red}{FLIR night}      & med     & high                   & low      & 1600x1800            & 1040 \\
    \textcolor{black}{RoadScene}          & high    & Aligned                    & Aligned      & variable          & 221 \\
    \hline
    \end{tabular}
    \label{tab:datasets}
\end{table}

\subsection{Experiment 1: Comparison of metrics for cross-modality registration quality}

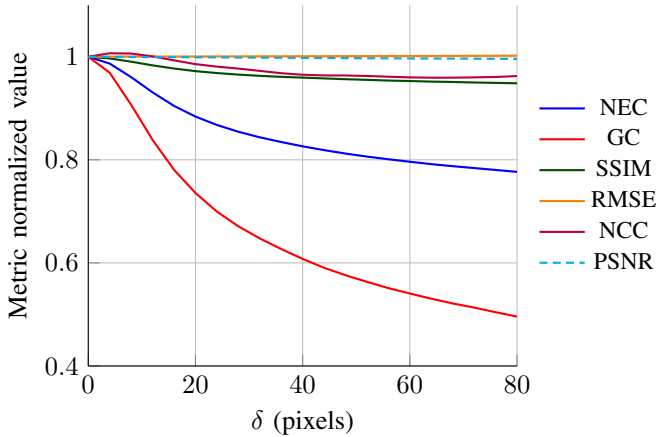
\begin{figure}[!h]
\centering

\begin{tikzpicture}

  \begin{axis}[
    axis y line*=left,
    axis x line*=bottom,
    xlabel={$\delta$ (pixels)},
    ylabel={Metric normalized value },
    grid=major,
    width=0.40\textwidth,
    height=0.35\textwidth,
    xmin=0, xmax=80,
    ymin=0.4, ymax=1.1,
    title={},
    legend style={
        at={(1.03,0.5)},
        anchor=west,
        draw=none,
        font=\small,
      },
  ]

  \addplot+[thick, mark=none, color=blue] table [x=Delta, y=NEC, col sep=space] {Graphs/metrics.dat};
  \addlegendentry{NEC}

  \addplot+[thick, mark=none, color=red] table [x=Delta, y=GC, col sep=space] {Graphs/metrics.dat};
  \addlegendentry{GC}

  \addplot+[thick, mark=none, color=green!60!black] table [x=Delta, y=SSIM, col sep=space] {Graphs/metrics.dat};
  \addlegendentry{SSIM}

  \addplot+[thick, mark=none, color=orange] table [x=Delta, y=RMSE, col sep=space] {Graphs/metrics.dat};
  \addlegendentry{RMSE}

  \addplot+[thick, mark=none, color=purple] table [x=Delta, y=NCC, col sep=space] {Graphs/metrics.dat};
  \addlegendentry{NCC}

  \addplot+[thick, mark=none, color=cyan] table [x=Delta, y=PSNR, col sep=space] {Graphs/metrics.dat};
  \addlegendentry{PSNR}

  \end{axis}

\end{tikzpicture}

\caption{Experiment 1: Metrics (divided by per v[$\delta=0$]) evolution with increasing disparity ($\delta$) between the infrared and visible registered images. We expect metrics suitable for cross-modality registration quality assessment to consistently decrease as the disparity increases.}
\label{fig:comparison_metrics}
\end{figure}

To compare the efficiency of different metrics to assess the quality of cross-modality registration, we take the RoadScene dataset~\cite{xu2020aaai}, which is already well aligned. 
\begin{itemize}
    \item \textbf{\acrshort{nec}}, our proposed metric.
    \item \textbf{\acrfull{gc}}~\cite{metric_gradient_correlation}, a metric based on the correlation of image gradients. Expected to work well for cross-modality registration.
    \item \textbf{\acrshort{ssim}}, The structural similarity should be maximized between the registered image and the reference image.
    \item \textbf{\acrshort{rmse}}, is expected to perform poorly for cross-modality registration as it's a purely intensity-based metric.
    \item \textbf{\acrshort{ncc}}, a classical metric for image registration, not optimized for cross-modality registration.
    \item \textbf{\acrshort{psnr}}, another classical metric for image quality, not meant for cross-modality registration.
\end{itemize}
For this experiment we will compute each compared metric on the 221 pairs of images adding 20 steps of increasing disparity (pure horizontal offset) between the visible and infrared images:
\small
\begin{equation}
    \begin{aligned}
        IR_{\delta}(u, v) &= IR(u+2\delta, v) \quad u \in [0, W-40] \\
        VIS_{\delta}(u, v) &= VIS(u-2\delta, v) \quad u \in [40, W] \\
        &\text{with} \quad v \in [0, H],\quad \delta \in [0, 20]
    \end{aligned}
\end{equation}
\normalsize
For each value of $\delta$, we compute the average value of each metric over the 221 image pairs. The results are reported in \cref{fig:comparison_metrics}. Since only \acrshort{nec} and \acrshort{gc} consistently decrease as the disparity increases, they are the only metrics that are fully suitable for assessing cross-modality registration quality. We also retain \acrshort{ssim} and \acrshort{ncc}, as they exhibit some sensitivity to misalignment, although this sensitivity is significantly weaker The remaining metrics, namely \acrshort{rmse} and \acrshort{psnr}, are not relevant for this task, as they do not show a consistent trend with increasing misalignment.
In the remainder of this work, we therefore use only \acrshort{nec}, \acrshort{gc}, \acrshort{ssim}, and \acrshort{ncc} as quality metrics for cross-modality registration.
\newcolumntype{M}[1]{>{\centering\arraybackslash}p{#1}}
\begin{table*}[!h]
    \caption{Experiment 2. Quantitative evaluation of image alignment. The \textit{Source} column reports the metric values computed on the images before registration. All other columns report the corresponding values after registration. The top three results for each metric are highlighted from first to third using \first{1.GREEN}, \second{2.BLUE}, and \third{3.CYAN}.}
    \centering
    \renewcommand{\arraystretch}{1.2}
    \setlength{\tabcolsep}{5pt}
    \begin{tabular}{M{1.8cm}lccccccc}
        \toprule
        \textbf{Dataset} & \textbf{Metric} & \textbf{Source} & \textbf{RIFT \cite{reg_homo_RIFT}} & \textbf{IMF \cite{reg_flow_IMF}} & \textbf{SuperFusion \cite{reg_flow_superfus}} & \textbf{CrossRAFT \cite{reg_flow_crossRaft}} & \textbf{PGMR \cite{zheng_plug-and-play_2025}} & \textbf{XCalib (ours)} \\
        \midrule
        \multirow{4}{*}{\makecell{\textit{Lynred day} \vspace{5mm}}}
& {\small NEC} $\uparrow$  & 0.45 & 0.54 (+21.2\%) & 0.46 (+2.8\%) & \first{0.60 (+34.3\%)} & \second{0.60 (+34.2\%)} & 0.56 (+26.8\%) & \third{0.57 (+28.8\%)} \\
& {\small GC} $\uparrow$  & 0.19 & 0.24 (+28.3\%) & 0.22 (+14.4\%) & \first{0.38 (+103.0\%)} & \second{0.35 (+87.0\%)} & \third{0.34 (+79.3\%)} & 0.33 (+74.1\%) \\
& {\small SSIM} $\uparrow$  & 0.48 & 0.50 (+4.2\%) & 0.49 (+2.6\%) & \second{0.52 (+7.7\%)} & \third{0.51 (+7.1\%)} & 0.51 (+6.6\%) & \first{0.52 (+7.9\%)} \\
& NCC $\uparrow$  & 0.14 & 0.15 (+12.3\%) & 0.14 (+4.5\%) & \first{0.16 (+18.0\%)} & \second{0.16 (+17.2\%)} & 0.15 (+14.3\%) & \second{0.16 (+17.2\%)} \\
\midrule
\multirow{4}{*}{\makecell{\textit{Lynred night} \vspace{5mm}}}
& {\small NEC} $\uparrow$  & 0.39 & 0.43 (+9.7\%) & 0.39 (+1.3\%) & \first{0.47 (+21.0\%)} & \second{0.46 (+19.1\%)} & 0.44 (+13.3\%) & \third{0.46 (+17.2\%)} \\
& {\small GC} $\uparrow$  & 0.05 & 0.07 (+33.7\%) & 0.06 (+31.6\%) & \third{0.12 (+145.0\%)} & 0.10 (+111.6\%) & \first{0.14 (+179.1\%)} & \second{0.13 (+173.1\%)} \\
& {\small SSIM} $\uparrow$  & 0.35 & 0.36 (+2.4\%) & 0.37 (+3.5\%) & \first{0.37 (+4.5\%)} & \first{0.37 (+4.5\%)} & \third{0.37 (+3.9\%)} & 0.36 (+0.6\%) \\
& NCC $\uparrow$  & 0.02 & 0.02 (+8.8\%) & 0.01 (-43.7\%) & 0.01 (-17.4\%) & \third{0.02 (+26.5\%)} & \second{0.02 (+28.9\%)} & \first{0.02 (+36.7\%)} \\
\midrule
\multirow{4}{*}{\makecell{\textit{FLIR day} \vspace{5mm}}}
& {\small NEC} $\uparrow$  & 0.37 & \second{0.53 (+45.6\%)} & 0.37 (+0.9\%) & \third{0.44 (+18.8\%)} & 0.40 (+8.1\%) & 0.42 (+14.3\%) & \first{0.56 (+52.8\%)} \\
& {\small GC} $\uparrow$  & 0.07 & \second{0.19 (+158.9\%)} & 0.10 (+27.4\%) & 0.15 (+103.1\%) & 0.07 (-11.4\%) & \third{0.16 (+114.8\%)} & \first{0.24 (+219.4\%)} \\
& {\small SSIM} $\uparrow$  & 0.40 & \second{0.45 (+14.0\%)} & \first{0.45 (+14.5\%)} & 0.44 (+9.9\%) & \third{0.45 (+13.0\%)} & 0.44 (+10.5\%) & 0.43 (+8.2\%) \\
& NCC $\uparrow$  & 0.60 & \second{0.70 (+15.6\%)} & 0.61 (+1.7\%) & 0.64 (+5.5\%) & 0.62 (+2.4\%) & \third{0.65 (+7.2\%)} & \first{0.70 (+15.9\%)} \\
\midrule
\multirow{4}{*}{\makecell{\textit{FLIR night} \vspace{5mm}}}
& {\small NEC} $\uparrow$  & 0.41 & 0.42 (+2.8\%) & 0.41 (+1.3\%) & \second{0.44 (+7.5\%)} & 0.39 (-3.1\%) & \third{0.43 (+4.6\%)} & \first{0.46 (+13.1\%)} \\
& {\small GC} $\uparrow$  & 0.06 & 0.08 (+28.5\%) & 0.09 (+38.8\%) & \third{0.09 (+49.7\%)} & 0.04 (-35.0\%) & \first{0.13 (+112.0\%)} & \second{0.11 (+81.5\%)} \\
& {\small SSIM} $\uparrow$  & 0.44 & 0.47 (+6.4\%) & \second{0.52 (+16.4\%)} & 0.48 (+8.8\%) & 0.47 (+6.8\%) & \third{0.49 (+10.0\%)} & \first{0.52 (+17.1\%)} \\
& NCC $\uparrow$  & 0.22 & \second{0.26 (+19.6\%)} & 0.22 (+1.2\%) & 0.23 (+6.0\%) & \first{0.26 (+20.5\%)} & 0.23 (+6.8\%) & \third{0.25 (+14.6\%)} \\
\midrule\midrule
\multirow{4}{*}{\makecell{\textbf{Average} \\ \textit{all datasets}}}
& {\small NEC} $\uparrow$  & 0.40 & \third{0.48 (+19.3\%)} & 0.41 (+1.6\%) & \second{0.49 (+20.8\%)} & 0.46 (+15.2\%) & 0.46 (+15.1\%) & \first{0.51 (+27.5\%)} \\
& {\small GC} $\uparrow$  & 0.09 & 0.15 (+55.0\%) & 0.12 (+23.3\%) & \third{0.19 (+99.7\%)} & 0.14 (+50.4\%) & \second{0.19 (+104.9\%)} & \first{0.20 (+117.2\%)} \\
& {\small SSIM} $\uparrow$  & 0.42 & 0.45 (+6.8\%) & \first{0.46 (+9.3\%)} & 0.45 (+7.8\%) & \third{0.45 (+7.9\%)} & \third{0.45 (+7.9\%)} & \second{0.45 (+8.9\%)} \\
& NCC $\uparrow$  & 0.24 & \second{0.28 (+15.9\%)} & 0.25 (+1.2\%) & 0.26 (+7.0\%) & \third{0.26 (+8.9\%)} & 0.26 (+8.5\%) & \first{0.28 (+16.2\%)} \\

    \end{tabular}

    \label{tab:quantitative_results}
\end{table*}

\subsection{Experiment 2. Frame to frame consistency.}  
\noindent

\begin{table}[H]
    \caption{Flow evaluation metrics. The best performances per metric and dataset is highlighted in \textcolor{green}{green}.}
    \centering
\begin{tabular}{l l c c c c}\toprule
Dataset & Method  & EPE$\downarrow$ & $<1$px$\uparrow$ & $<3$px$\uparrow$ & $<5$px$\uparrow$ \\ \midrule
\multirow{5}{*}{\makecell{Lynred\\day}} & SFusion  & 0.86 & 0.75 & 0.95 & 0.98 \\ 
& RIFT  & 5.19 & 0.28 & 0.69 & 0.83 \\ 
& C-RAFT  & 0.70 & 0.83 & 0.97 & 0.99 \\ 
& PGMR  & 0.85 & 0.74 & 0.97 & 0.99 \\ 
& Ours  & \textbf{\textcolor{green}{0.53}} & \textbf{\textcolor{green}{0.90}} & \textbf{\textcolor{green}{0.98}} & \textbf{\textcolor{green}{0.99}} \\ 
\midrule
\multirow{5}{*}{\makecell{Lynred\\night}} & SFusion  & 1.68 & 0.51 & 0.84 & 0.94 \\ 
& RIFT  & 5.14 & 0.11 & 0.46 & 0.69 \\ 
& C-RAFT  & 1.41 & 0.58 & 0.89 & 0.96 \\ 
& PGMR  & 1.68 & 0.42 & 0.87 & 0.95 \\ 
& Ours  & \textbf{\textcolor{green}{0.94}} & \textbf{\textcolor{green}{0.77}} & \textbf{\textcolor{green}{0.94}} & \textbf{\textcolor{green}{0.97}} \\ 
\midrule
\multirow{5}{*}{\makecell{FLIR\\night}} & SFusion  & 10.69 & 0.07 & 0.28 & 0.44 \\ 
& RIFT  & 14.34 & 0.03 & 0.17 & 0.31 \\ 
& C-RAFT  & 16.12 & 0.11 & 0.34 & 0.46 \\ 
& PGMR  & 9.17 & 0.07 & 0.34 & 0.53 \\ 
& Ours  & \textbf{\textcolor{green}{8.44}} & \textbf{\textcolor{green}{0.23}} & \textbf{\textcolor{green}{0.55}} & \textbf{\textcolor{green}{0.67}} \\ 
\midrule
\multirow{5}{*}{\makecell{FLIR\\day}} & SFusion  & 19.91 & 0.06 & 0.20 & 0.32 \\ 
& RIFT  & 12.12 & 0.07 & 0.36 & 0.55 \\ 
& C-RAFT  & 15.89 & 0.19 & 0.44 & 0.54 \\ 
& PGMR  & 16.61 & 0.04 & 0.21 & 0.35 \\ 
& Ours  & \textbf{\textcolor{green}{10.10}} & \textbf{\textcolor{green}{0.28}} & \textbf{\textcolor{green}{0.59}} & \textbf{\textcolor{green}{0.68}} \\ 
\bottomrule
\end{tabular}
\label{tab:flow_eval}
\end{table}
 Another crucial aspect is the stability of the produced registration for each method. Since our experiments use video sequences, we compute the frame-to-frame optical flow separately for the visible and registered infrared channels. The flow is estimated using a pretrained RAFT~\cite{raft_flow} model, with the visible flow serving as the reference. We compute the L2 norm between the two flows to quantify stability. For this experiment, only the four best-performing methods were evaluated on the sequential FLIR~\cite{flirDataset} and Lynred Mobility datasets~\cite{lynred_lynred_2025}. The results are summarized in \cref{tab:flow_eval}.\\
Our method benefits from the consistency of metric depth estimation, resulting in smooth frame-to-frame alignment without local deformations. This is the primary limitation of keypoint-based methods like RIFT: the estimated homography varies from frame to frame, making the registration flow unstable.
 FLIR \acrfull{epe} is higher because the flow between frame is much higher due to the low framerate.

\subsection{Experiment 3: Comparison with \acrshort{sota} methods}

The objective is for each dataset to register for all frames the infrared image to the corresponding visible image.
All chosen comparison methods provide pretrained weights, except for CrossRAFT.\\
To train CrossRAFT~\cite{reg_flow_crossRaft}, we followed the procedure outlined in the original paper, training the network over 10k steps using the YouTube-VOS dataset.

\textbf{Our method.}
We use the training strategy and parameters described in Section~\ref{sec:methodology}. Camera calibration takes approximately one minute on an NVIDIA\textregistered{} A5000 GPU, followed by image registration at approximately 10~images/s using Depth Anything V2~\cite{depth_anything_v2}. No post-processing is applied to occlusion-induced artifacts, as the objective is to evaluate the registration method itself.

\newlength{\tablength}
\setlength{\tablength}{0.165\textwidth}
\begin{figure*}[hb!]
    \centering
    \renewcommand{\arraystretch}{1} 
    \begin{tabularx}{\textwidth}{ c @{\hspace{0pt}} c @{\hspace{0pt}} c @{\hspace{0pt}} c @{\hspace{0pt}} c @{\hspace{0pt}} c @{\hspace{0pt}} c @{\hspace{0pt}}}

        \textbf{Source}  & \textbf{RIFT} & \textbf{SuperFusion} & \textbf{CrossRAFT}  & \textbf{PGMR} & \textbf{XCalib (ours)} \\

        \includegraphics[width=\tablength]{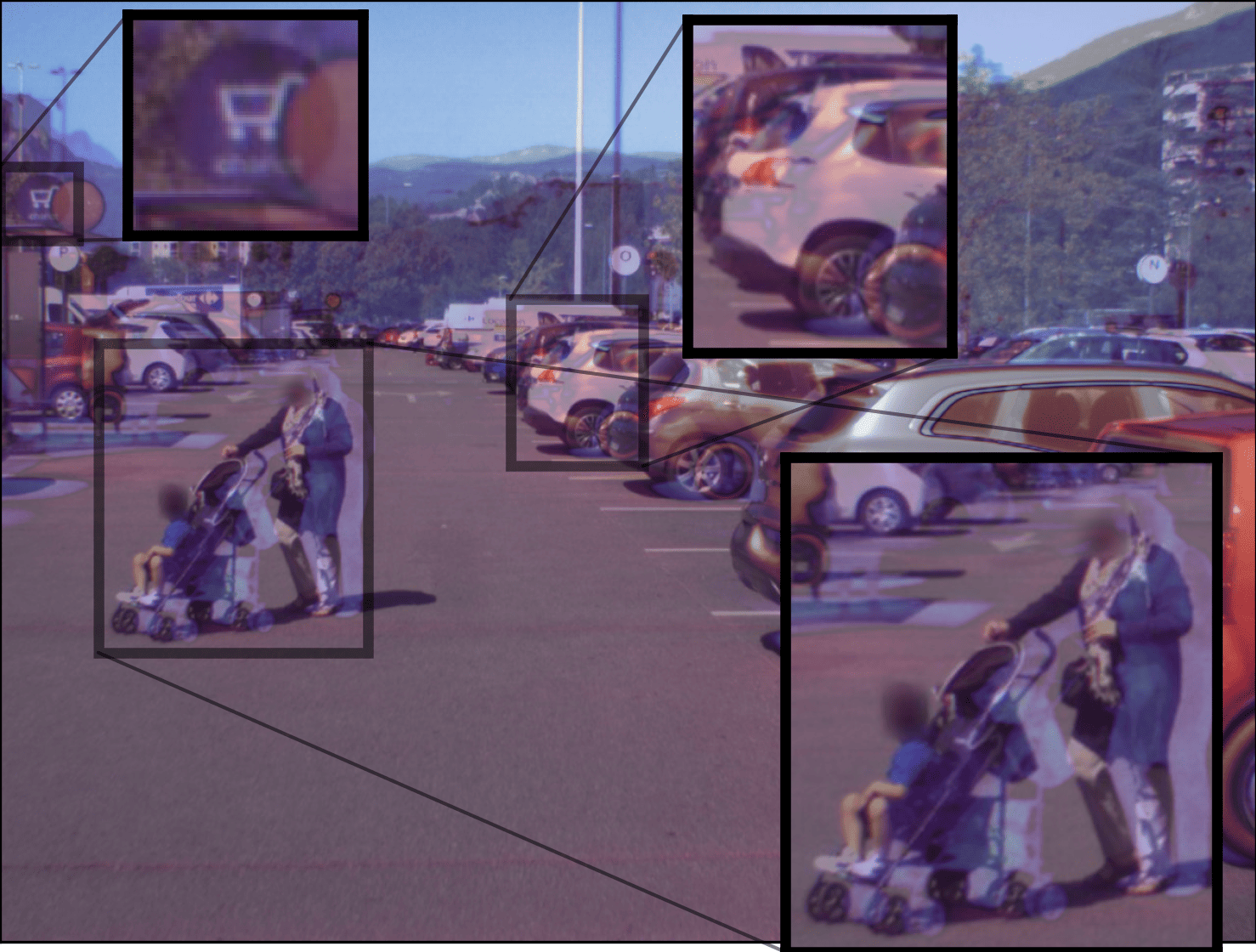} &
        \includegraphics[width=\tablength]{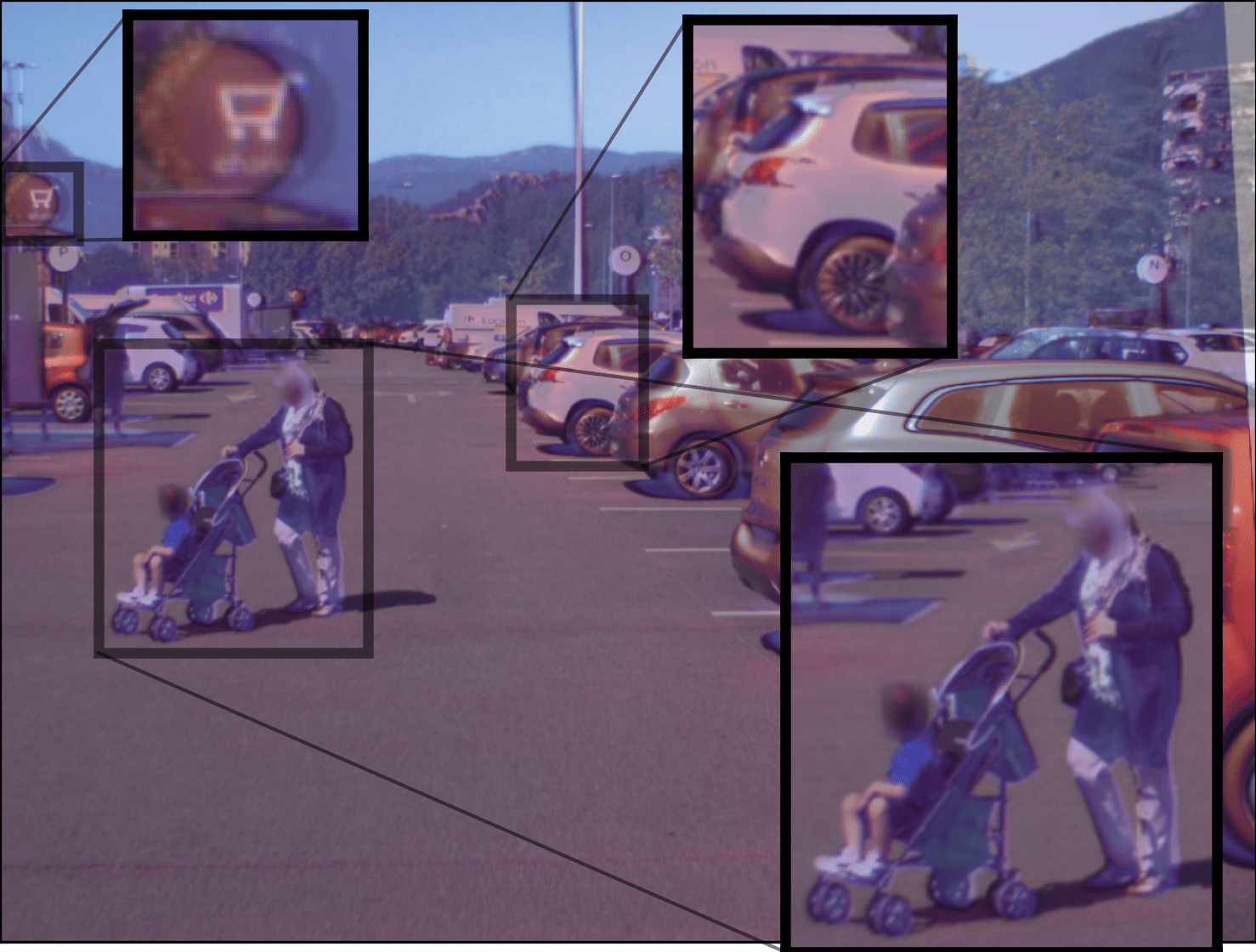} &
        \includegraphics[width=\tablength]{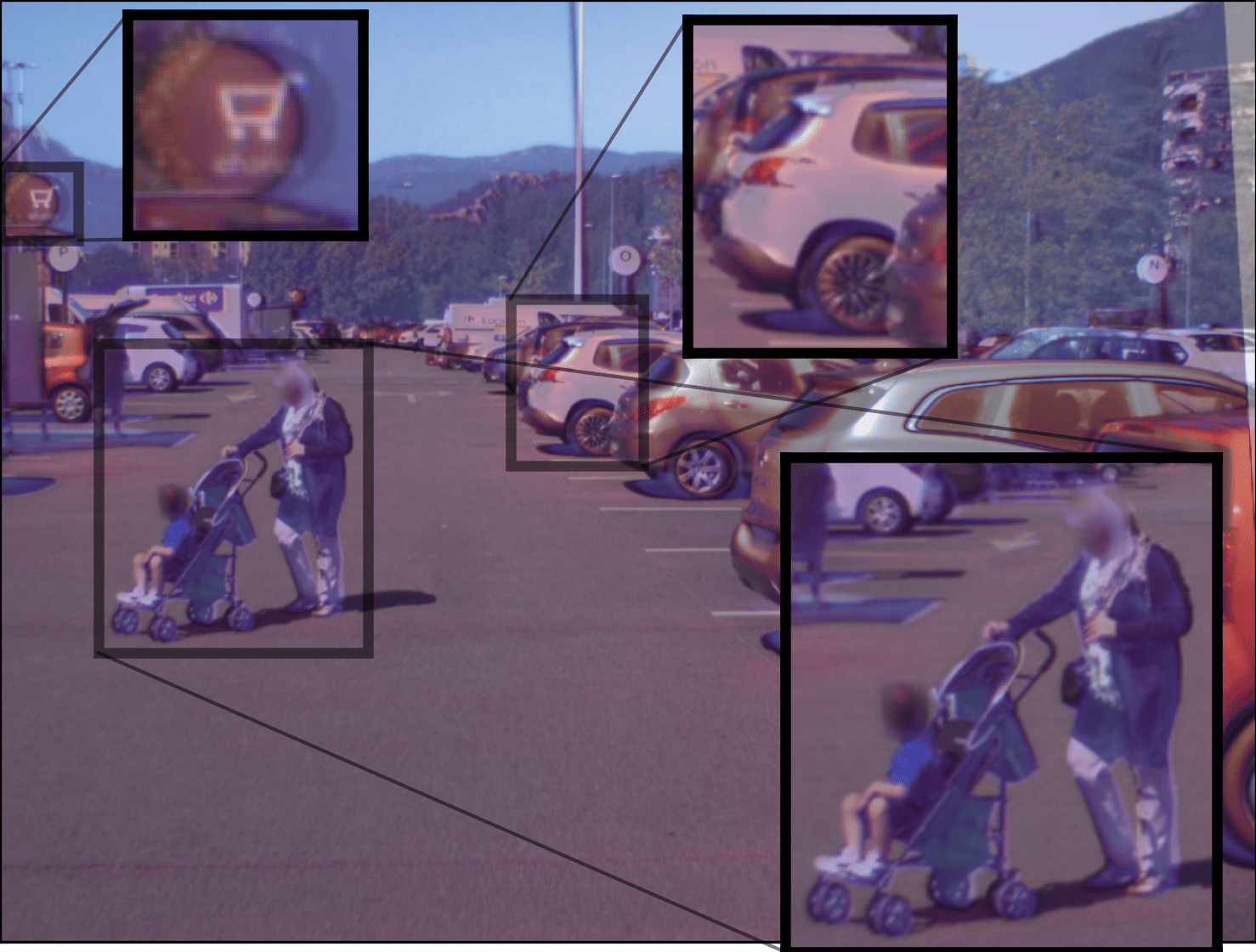} &
        \includegraphics[width=\tablength]{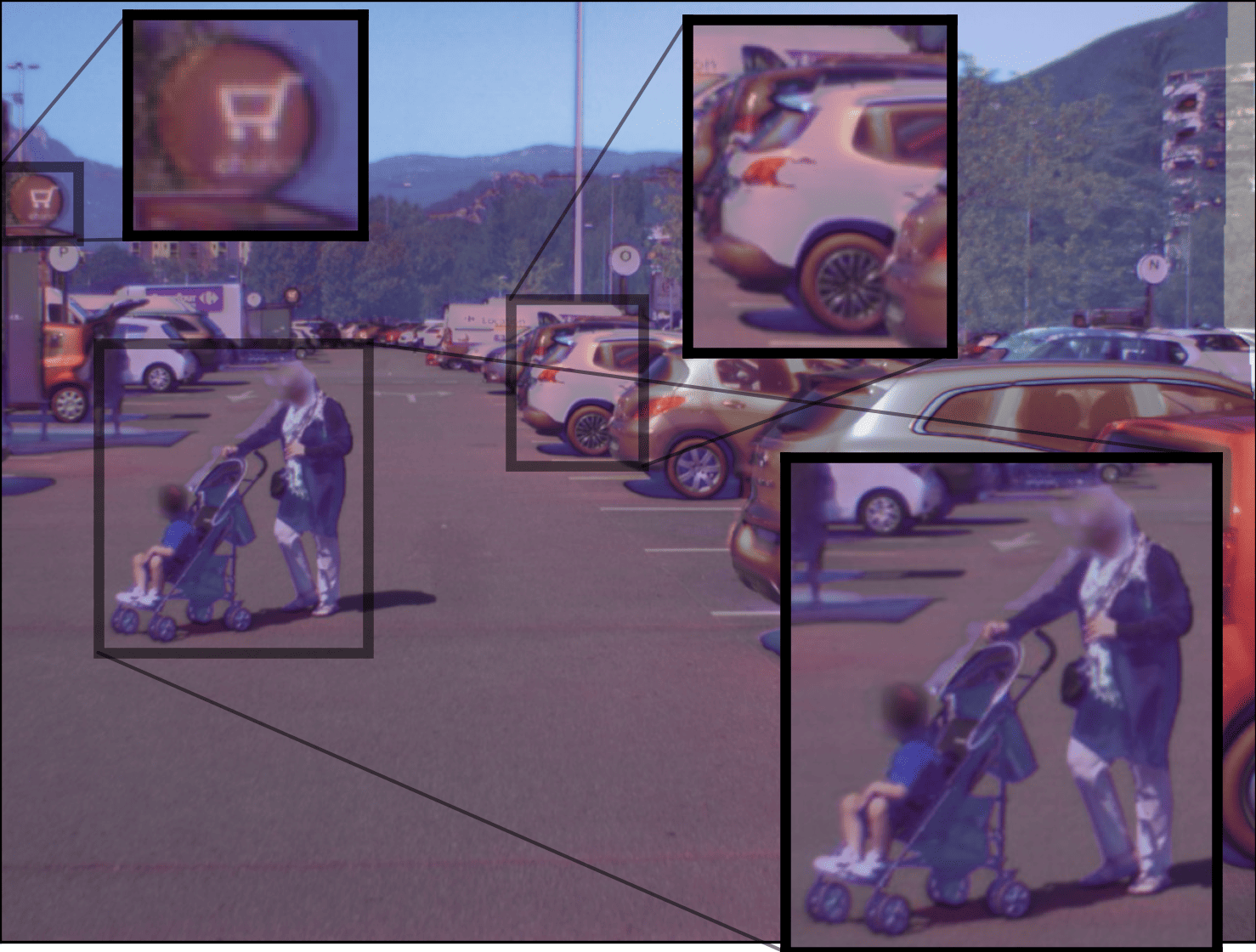} &
        \includegraphics[width=\tablength]{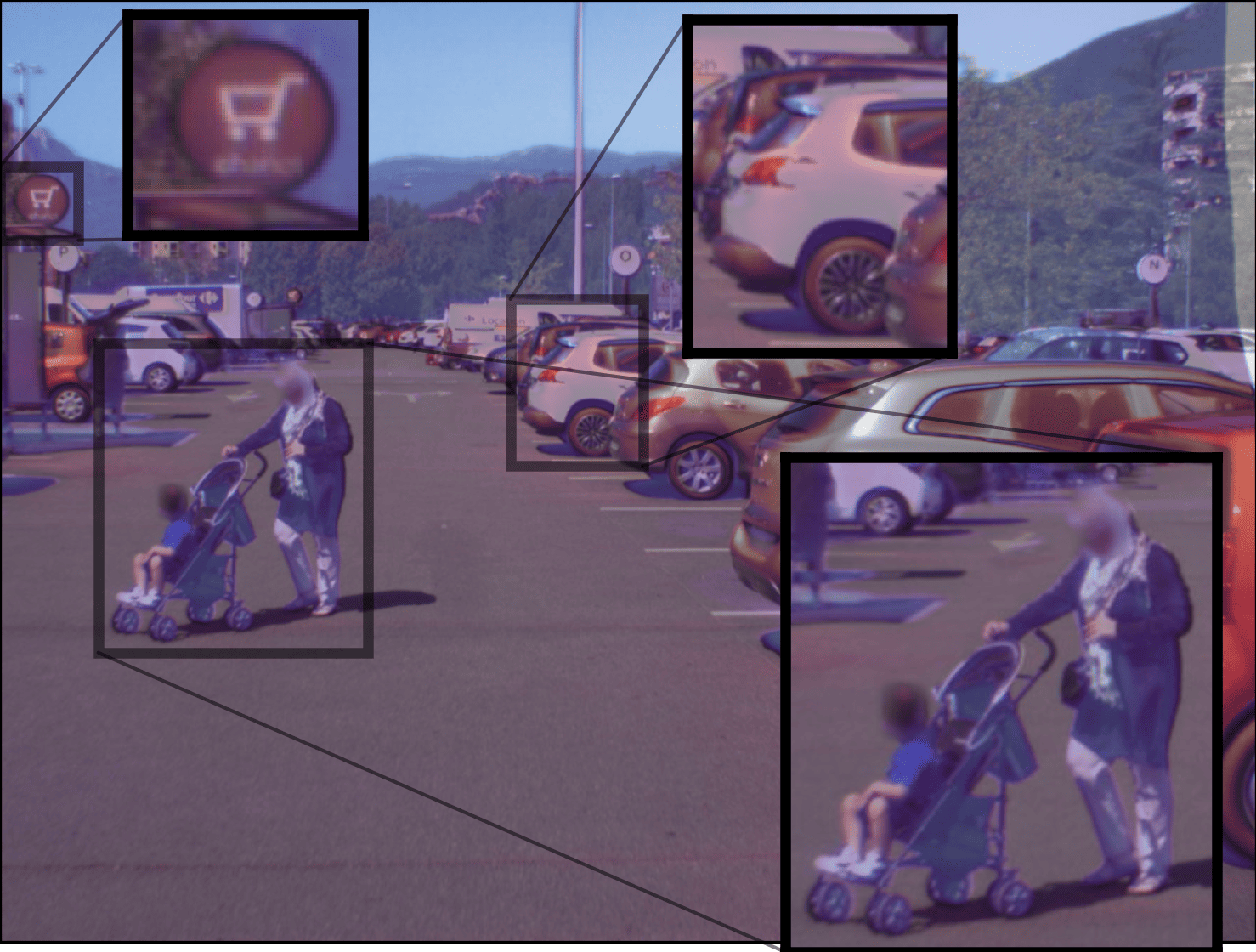} &
        \includegraphics[width=\tablength]{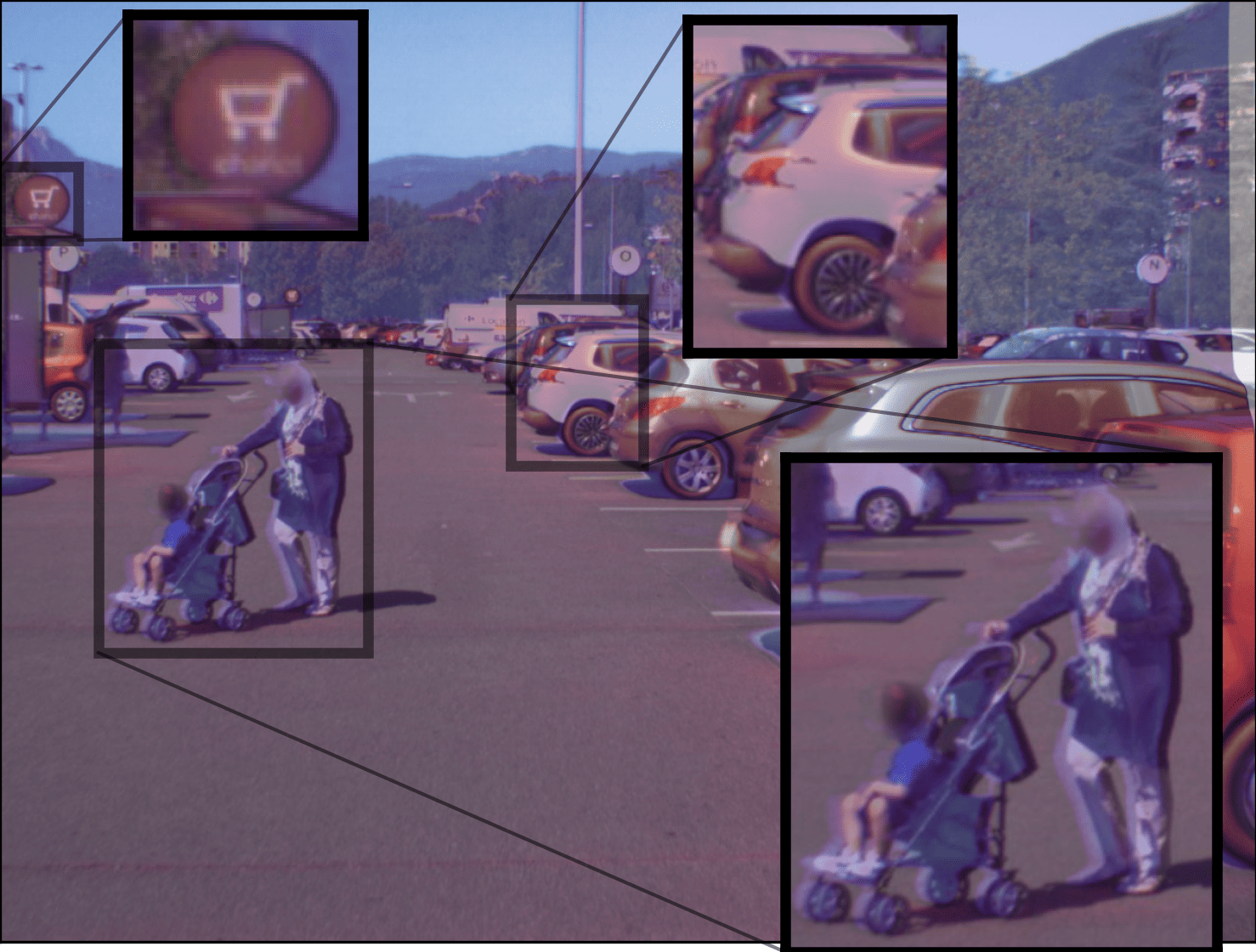} \\

        \includegraphics[width=\tablength]{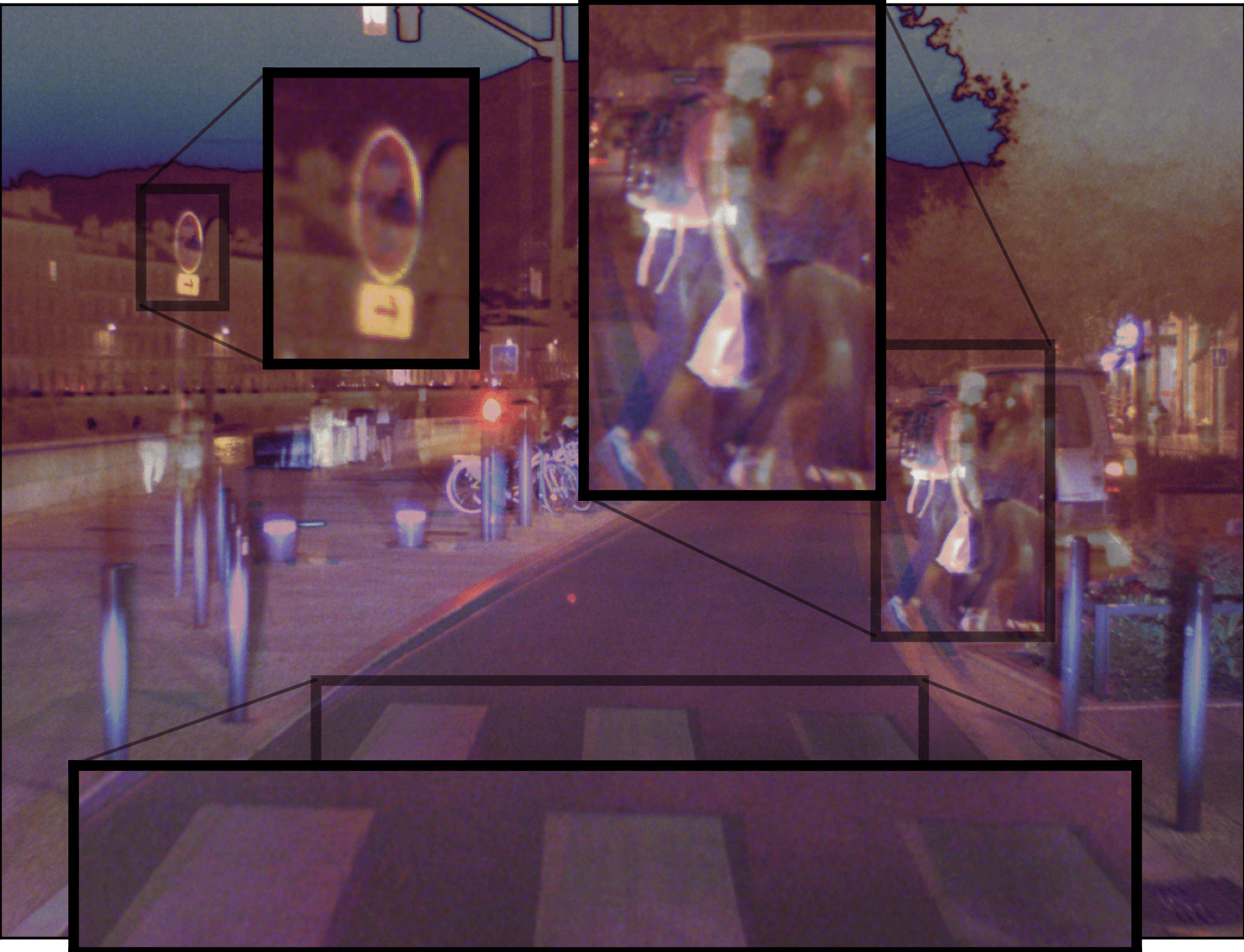} &
        \includegraphics[width=\tablength]{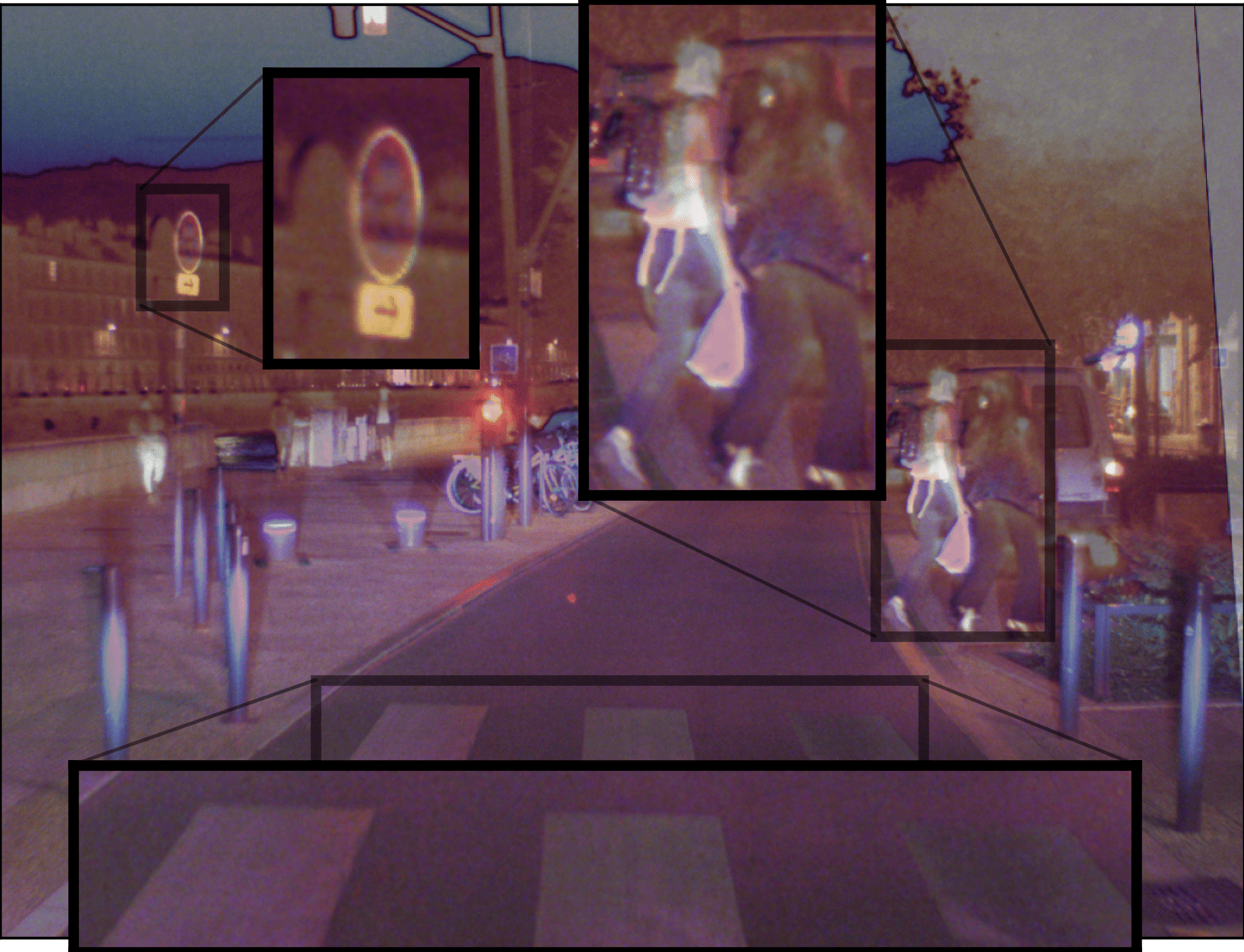} &
        \includegraphics[width=\tablength]{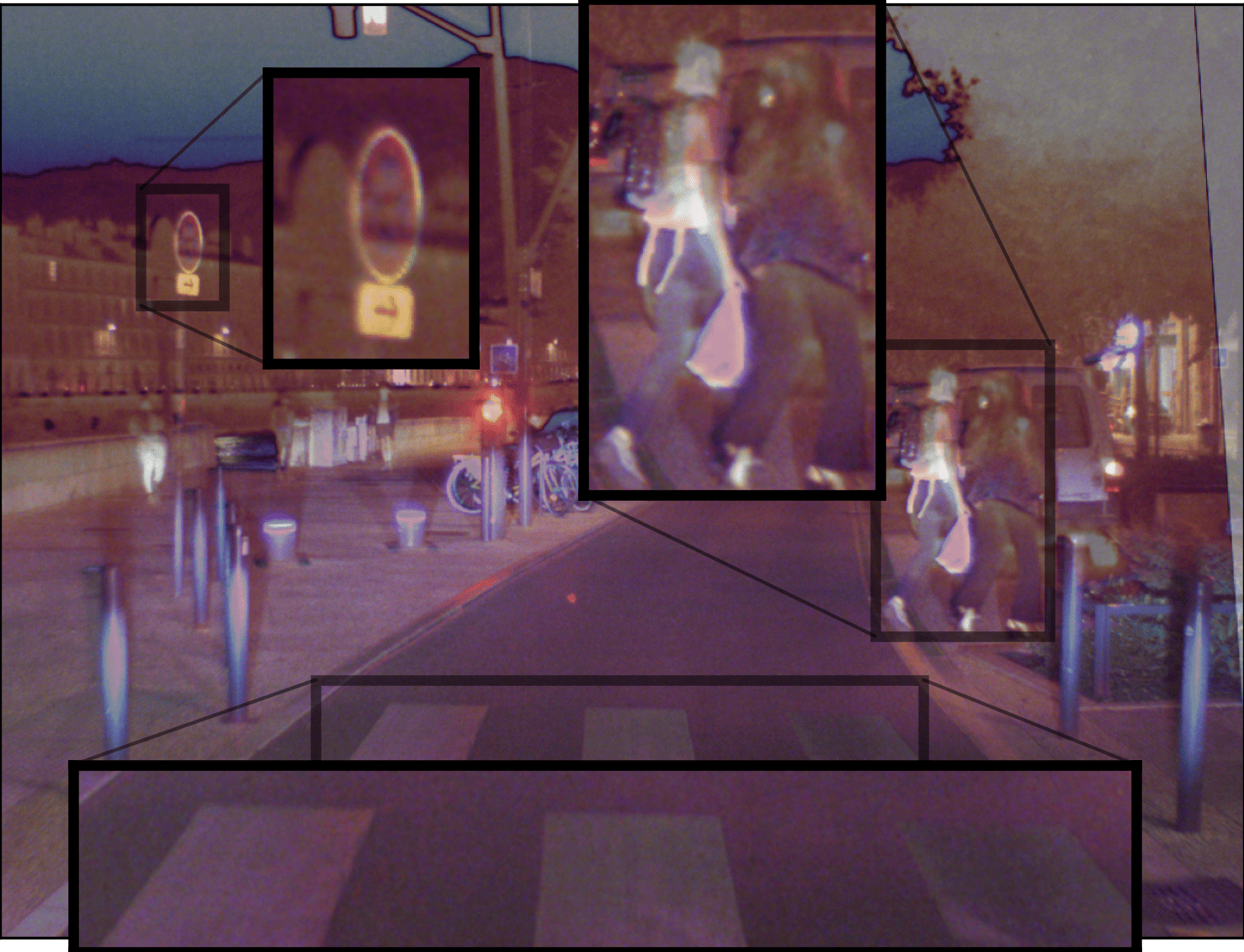} &
        \includegraphics[width=\tablength]{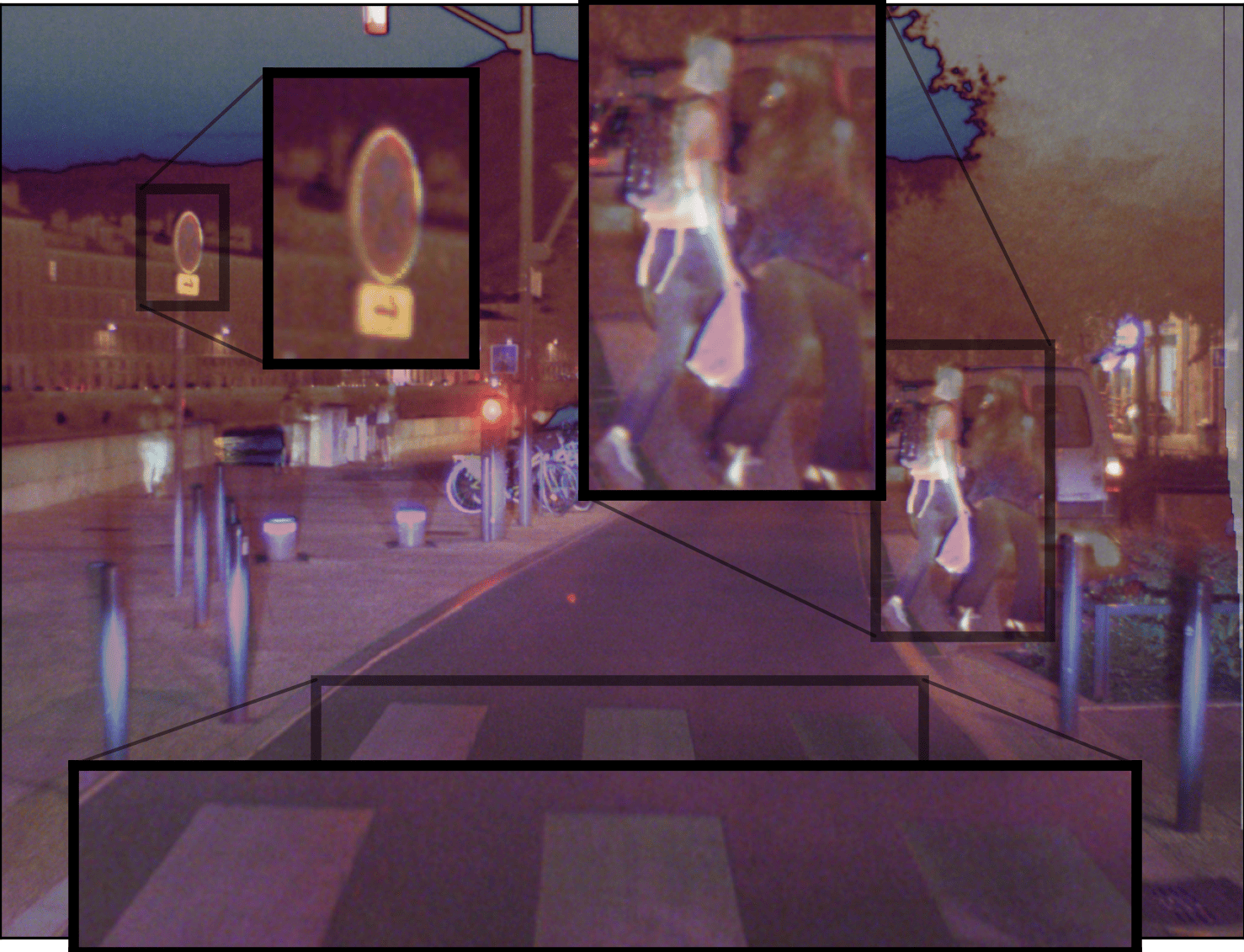} &
        \includegraphics[width=\tablength]{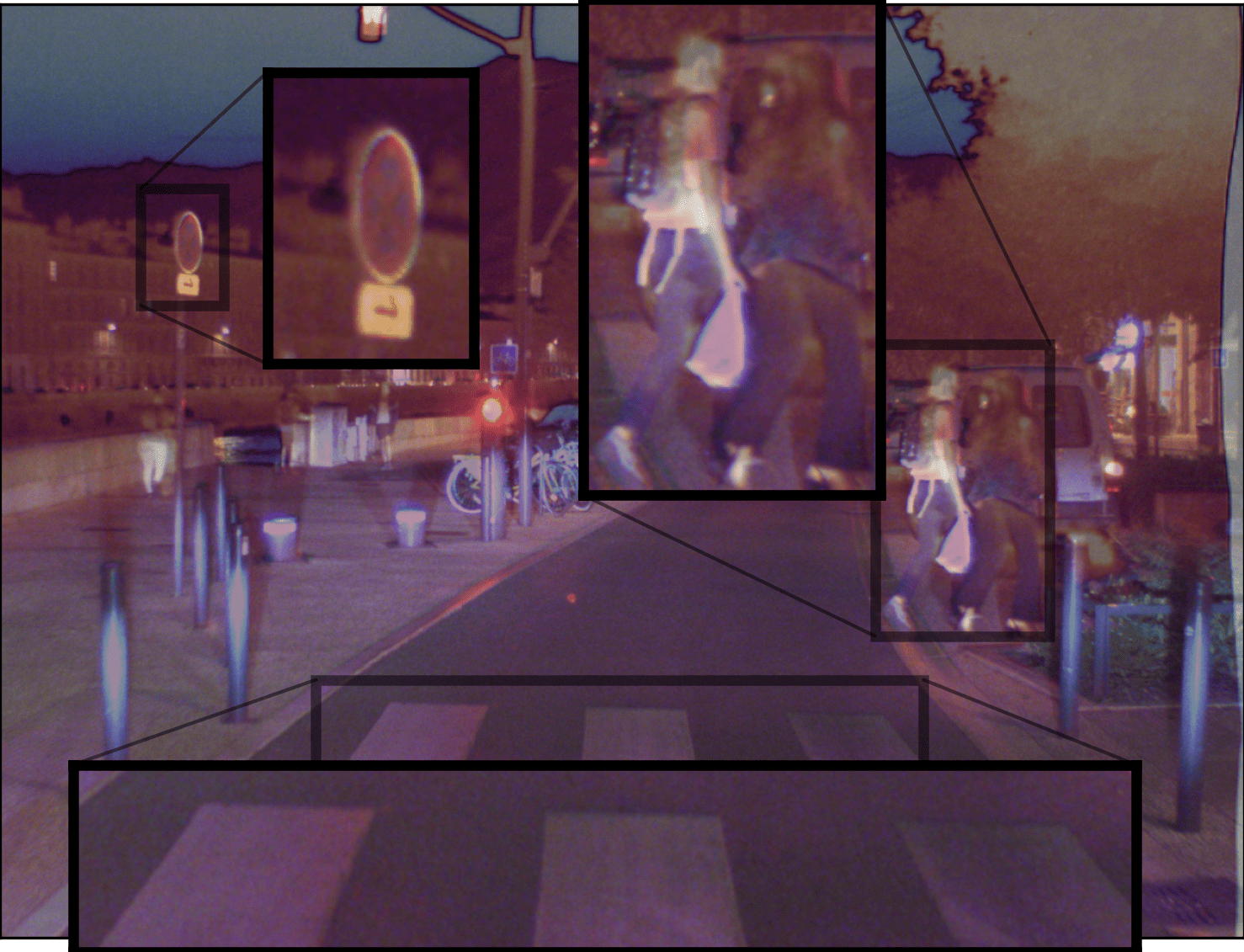} &
        \includegraphics[width=\tablength]{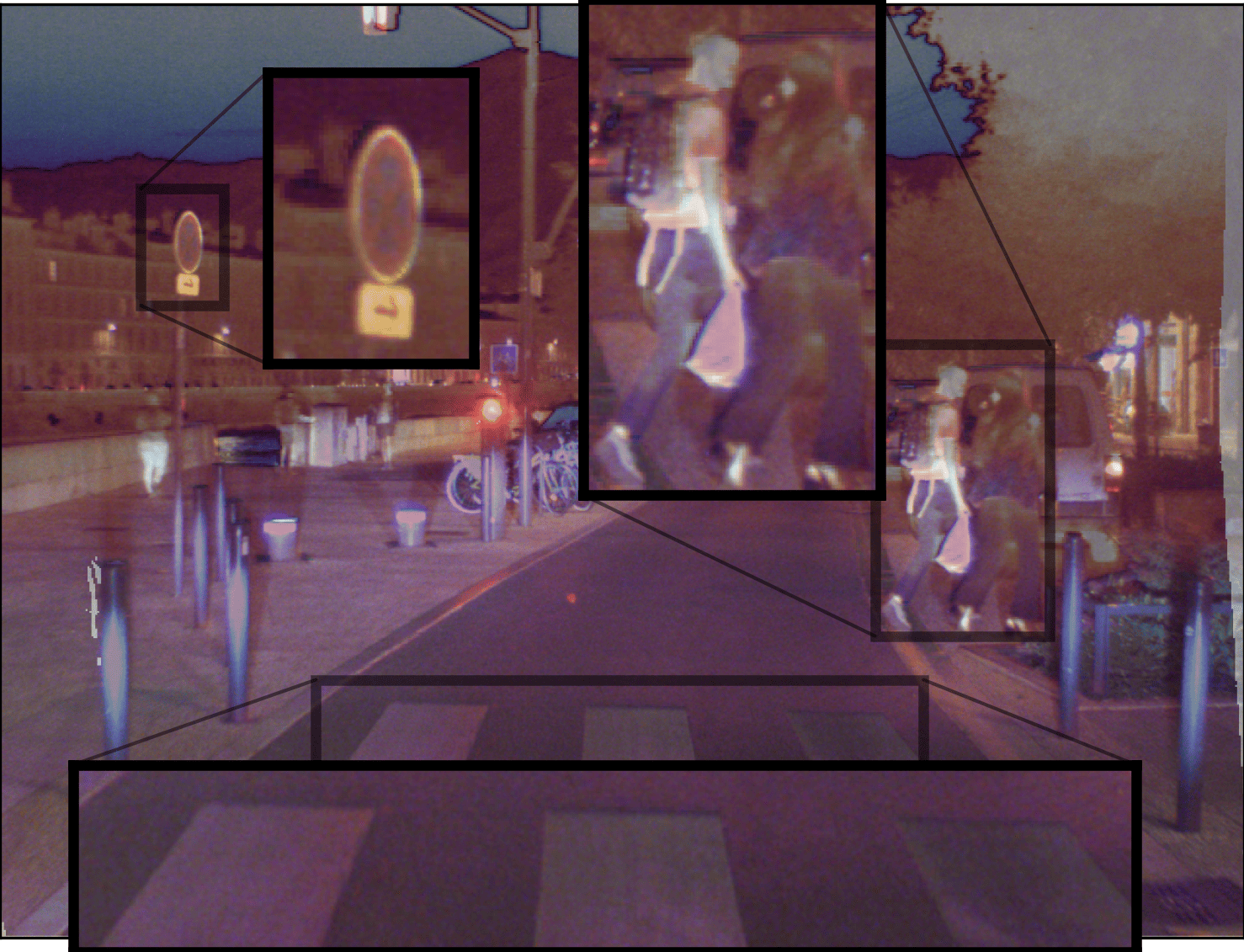} \\

        \includegraphics[width=\tablength]{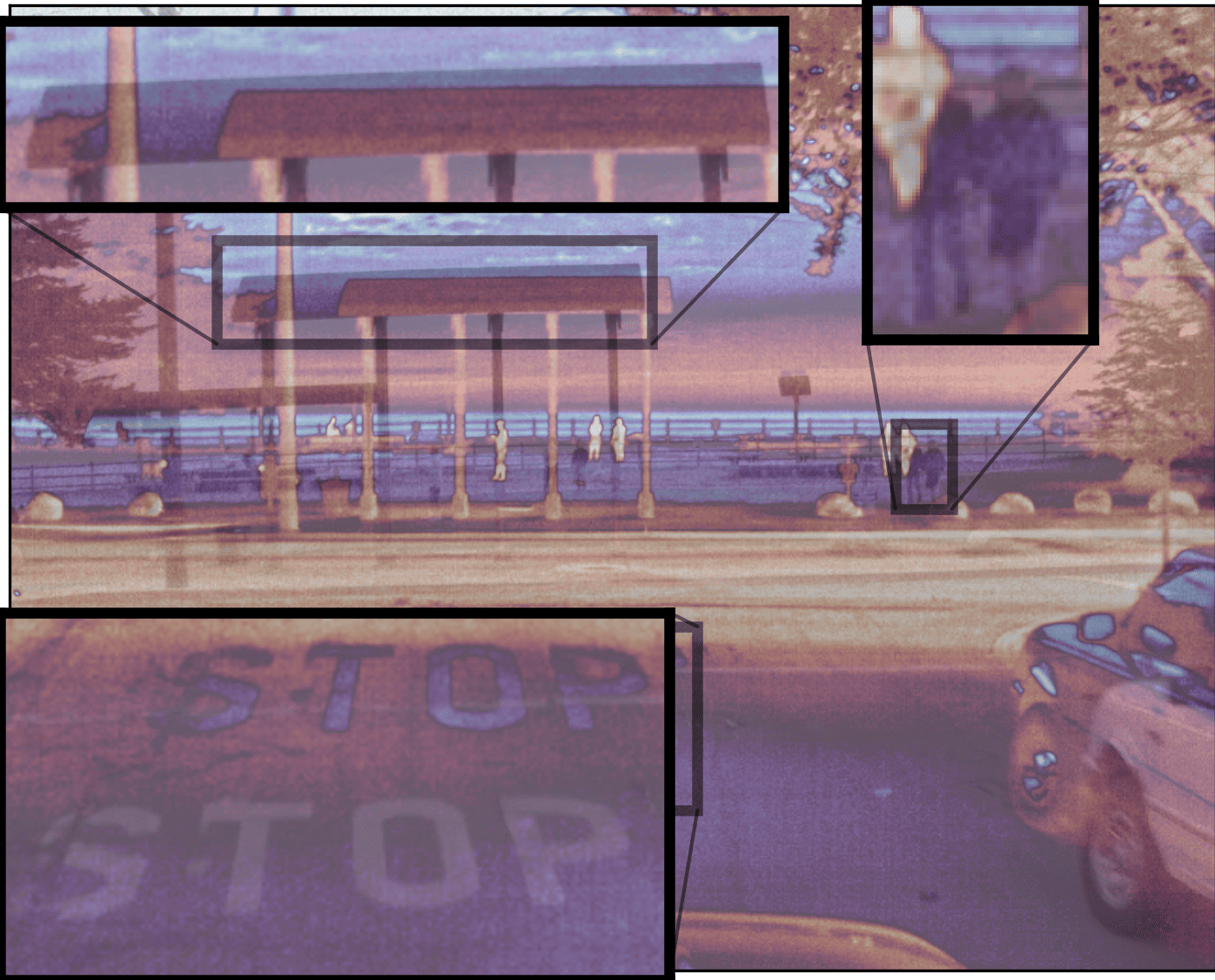} &
        \includegraphics[width=\tablength]{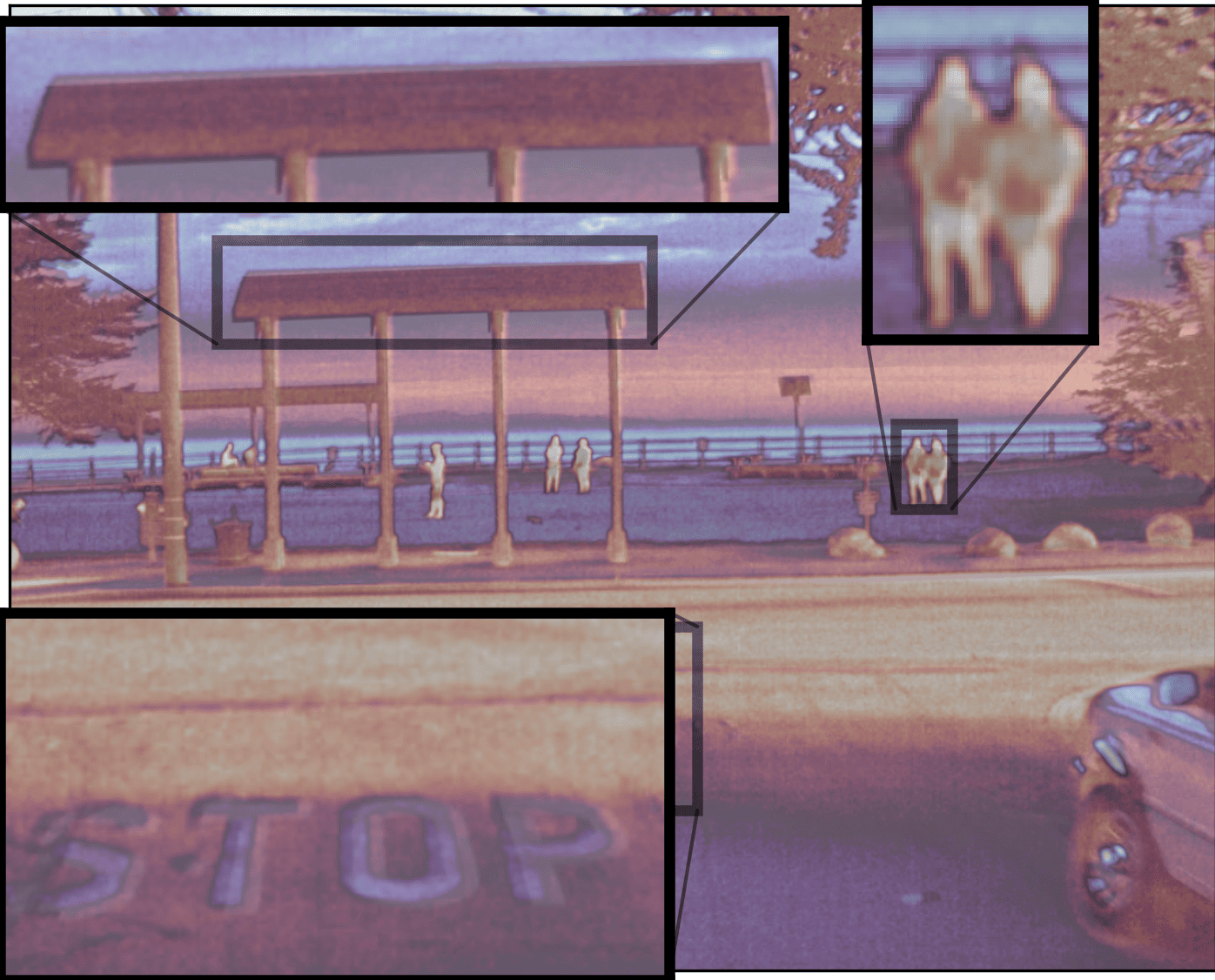} &
        \includegraphics[width=\tablength]{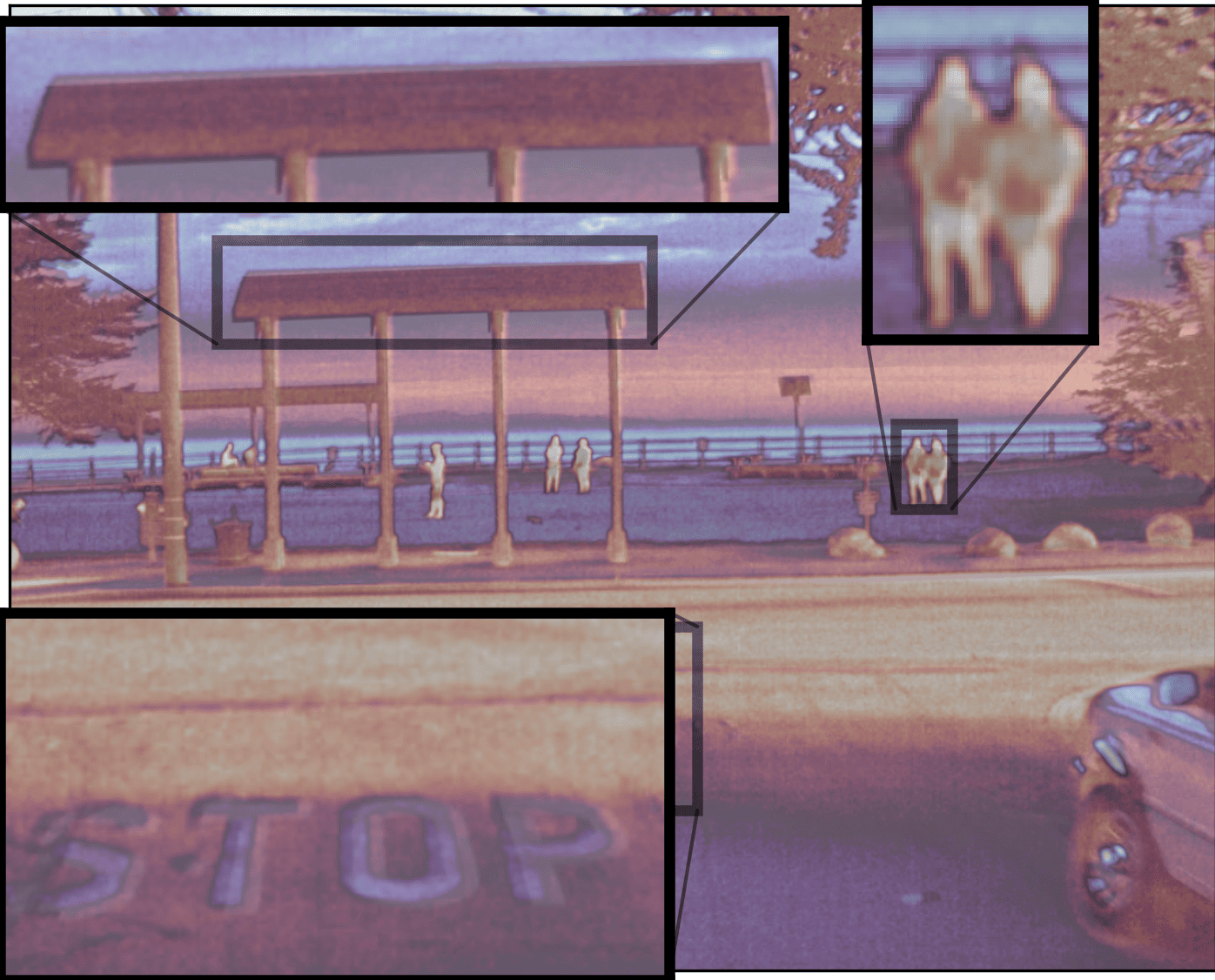} &
        \includegraphics[width=\tablength]{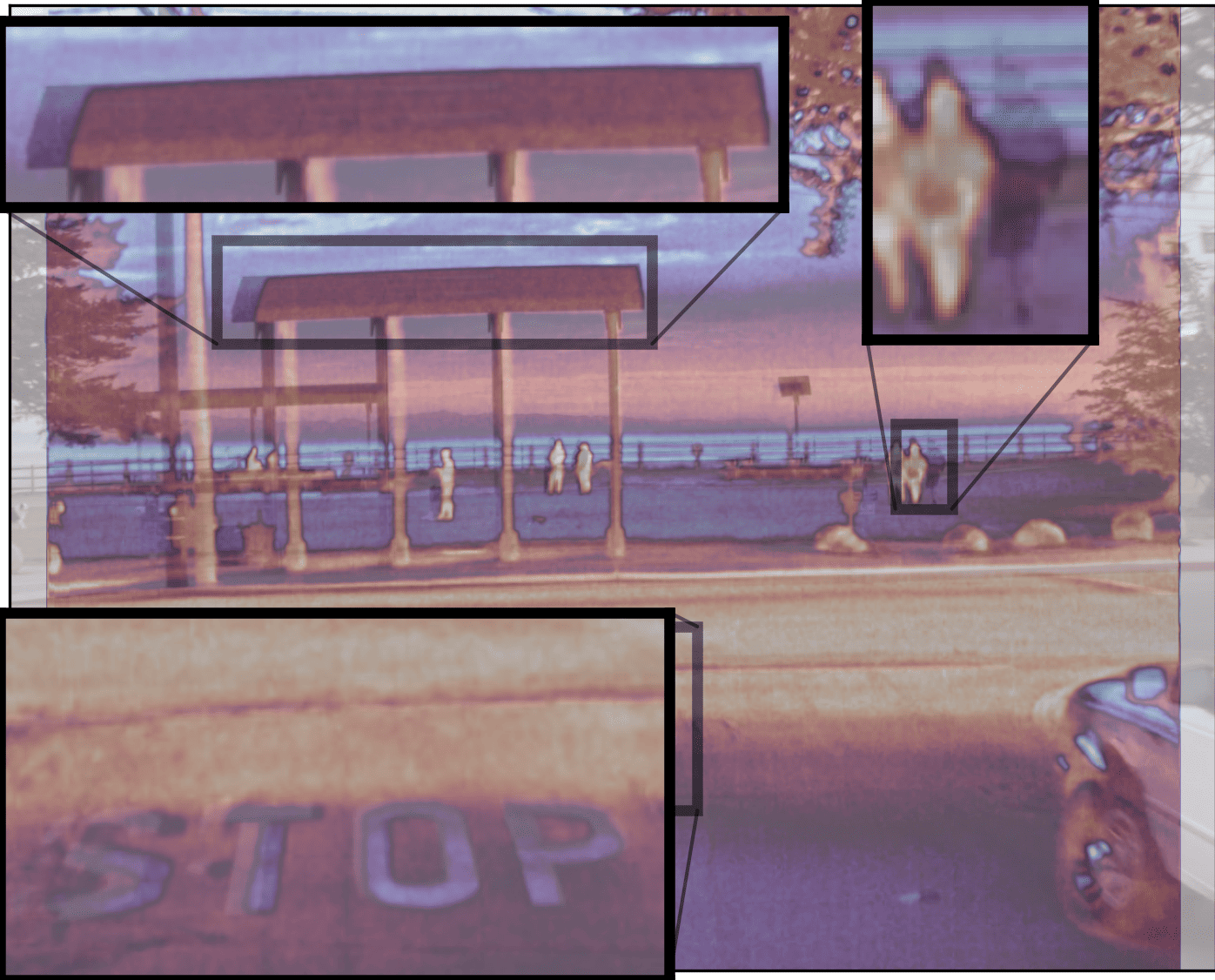} &
        \includegraphics[width=\tablength]{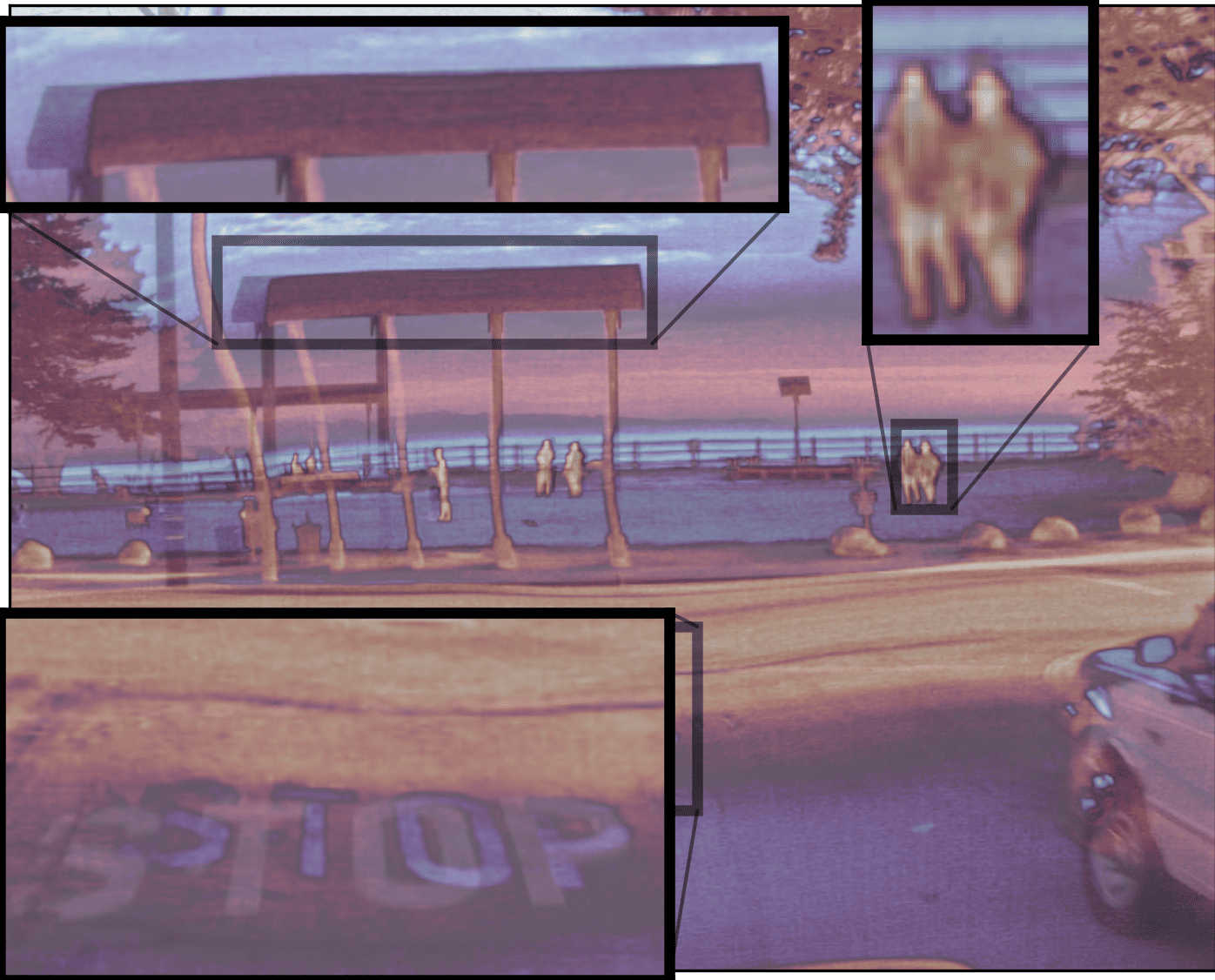} &
        \includegraphics[width=\tablength]{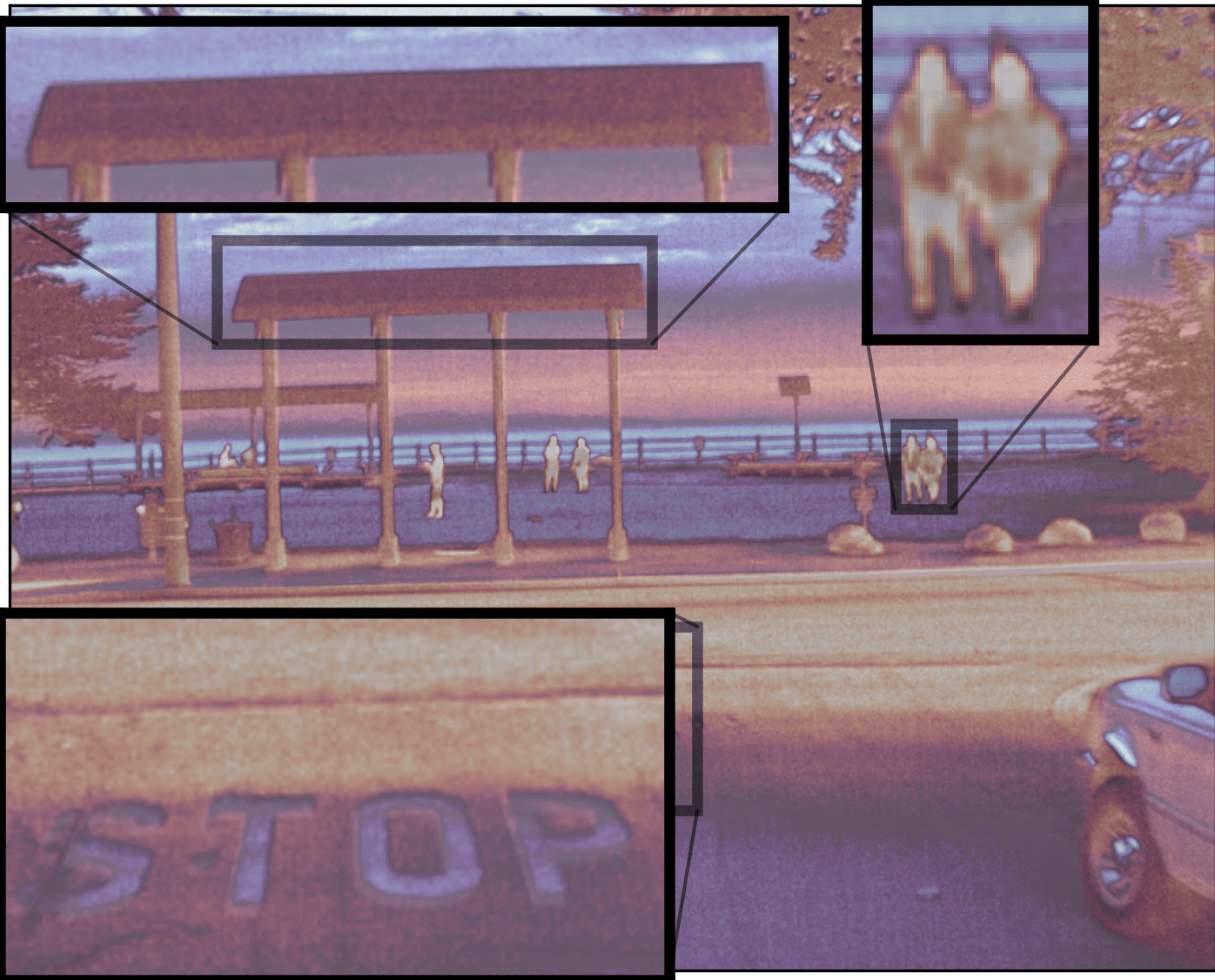} \\

        \includegraphics[width=\tablength]{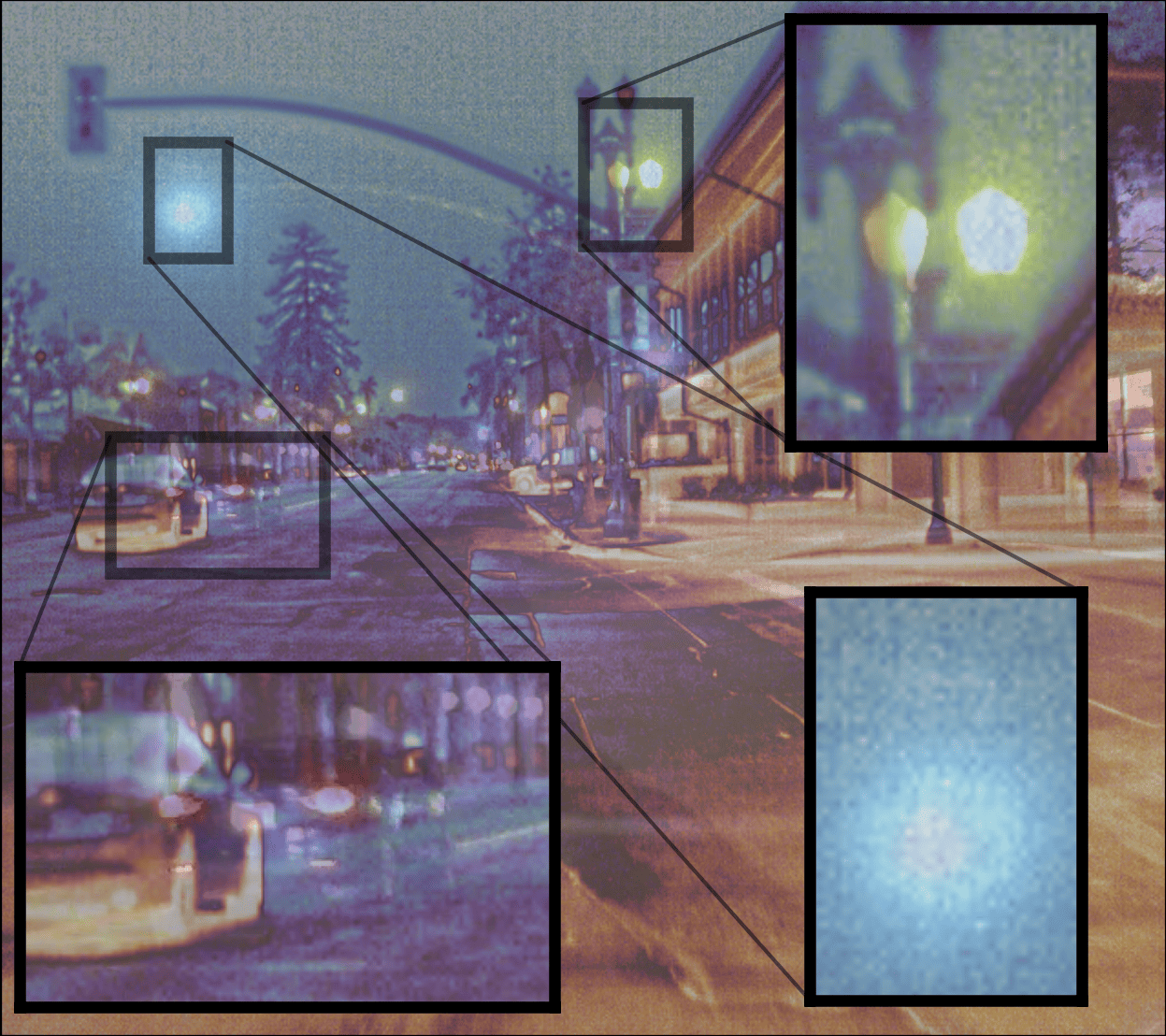} &
        \includegraphics[width=\tablength]{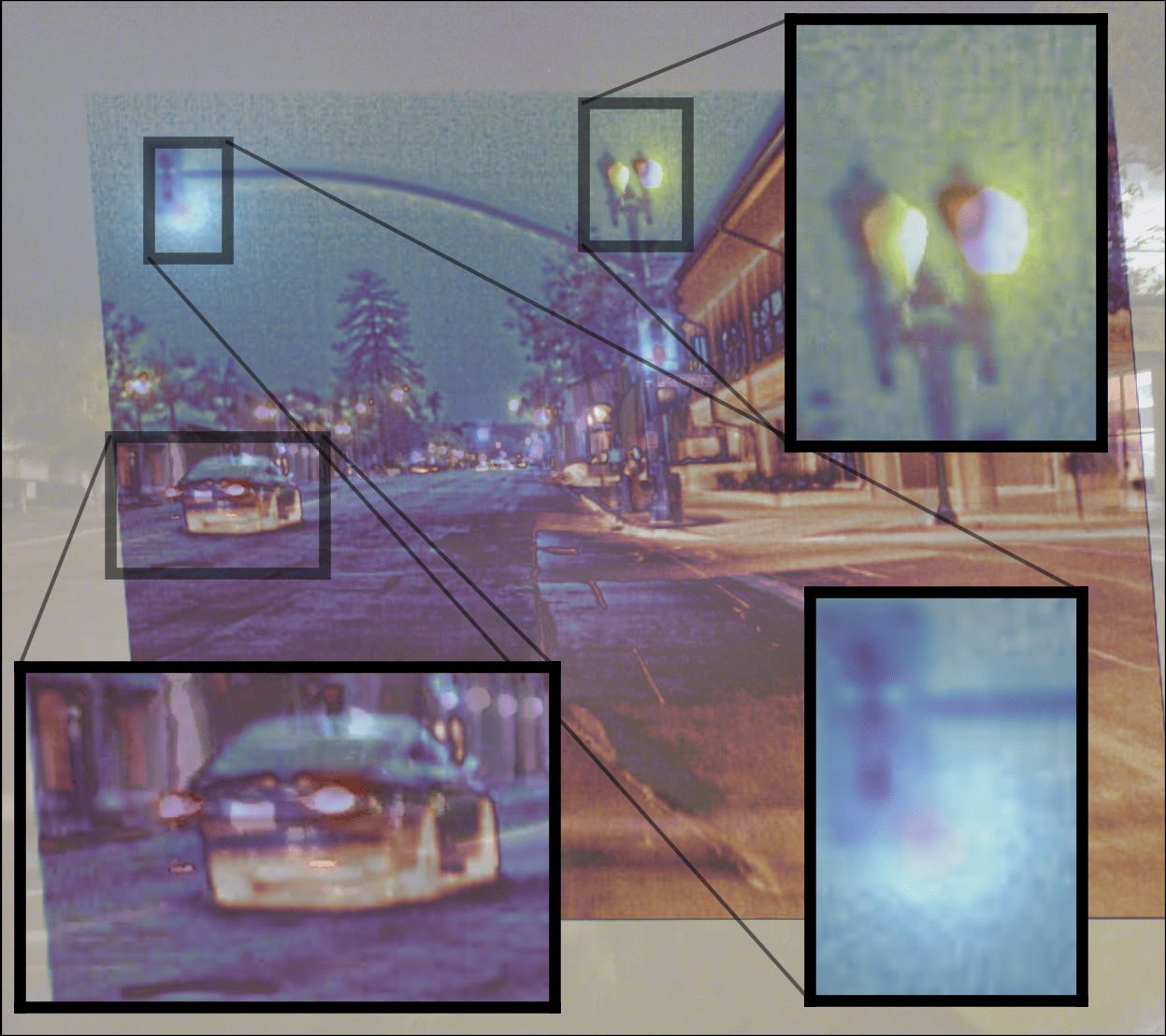} &
        \includegraphics[width=\tablength]{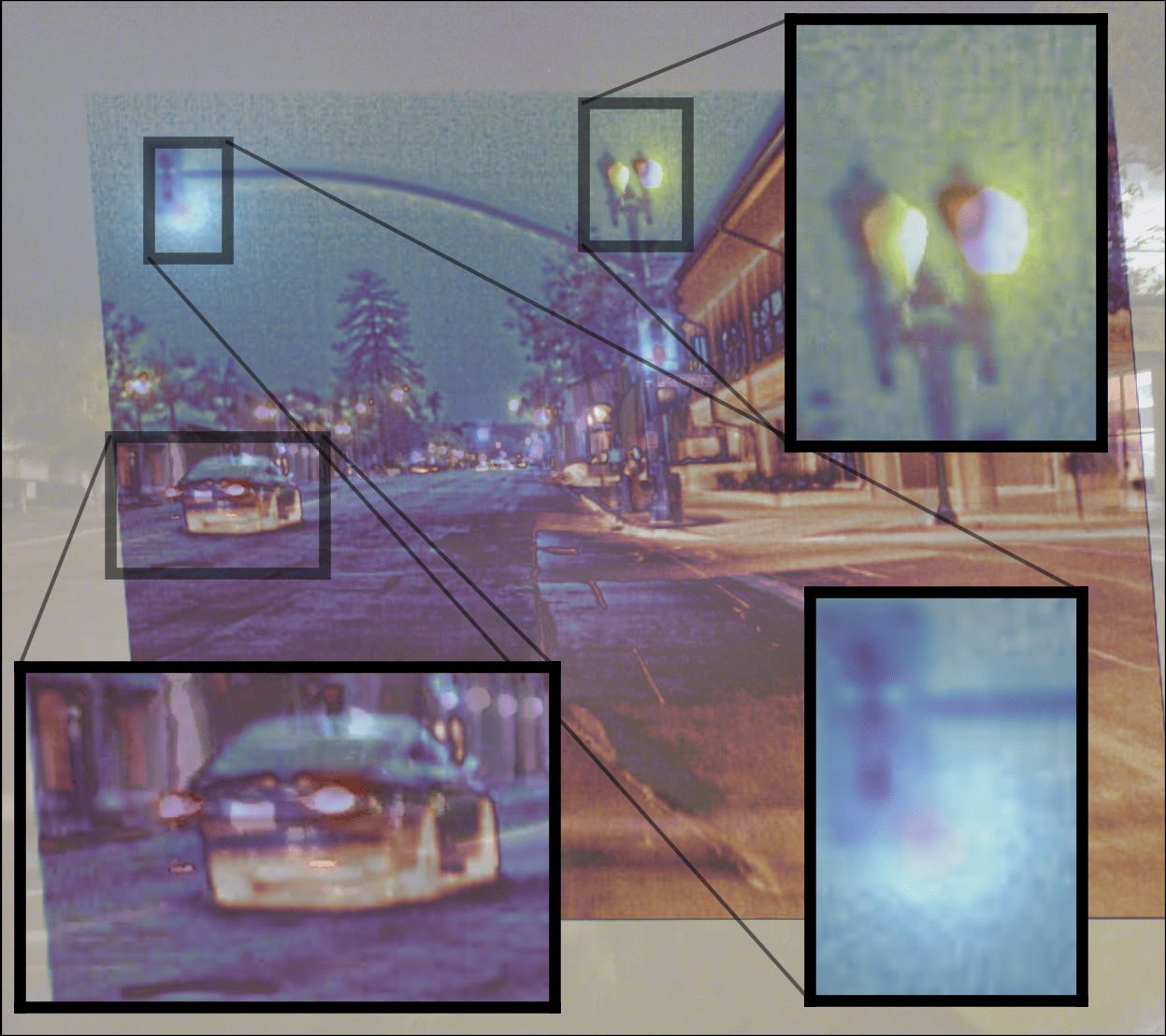} &
        \includegraphics[width=\tablength]{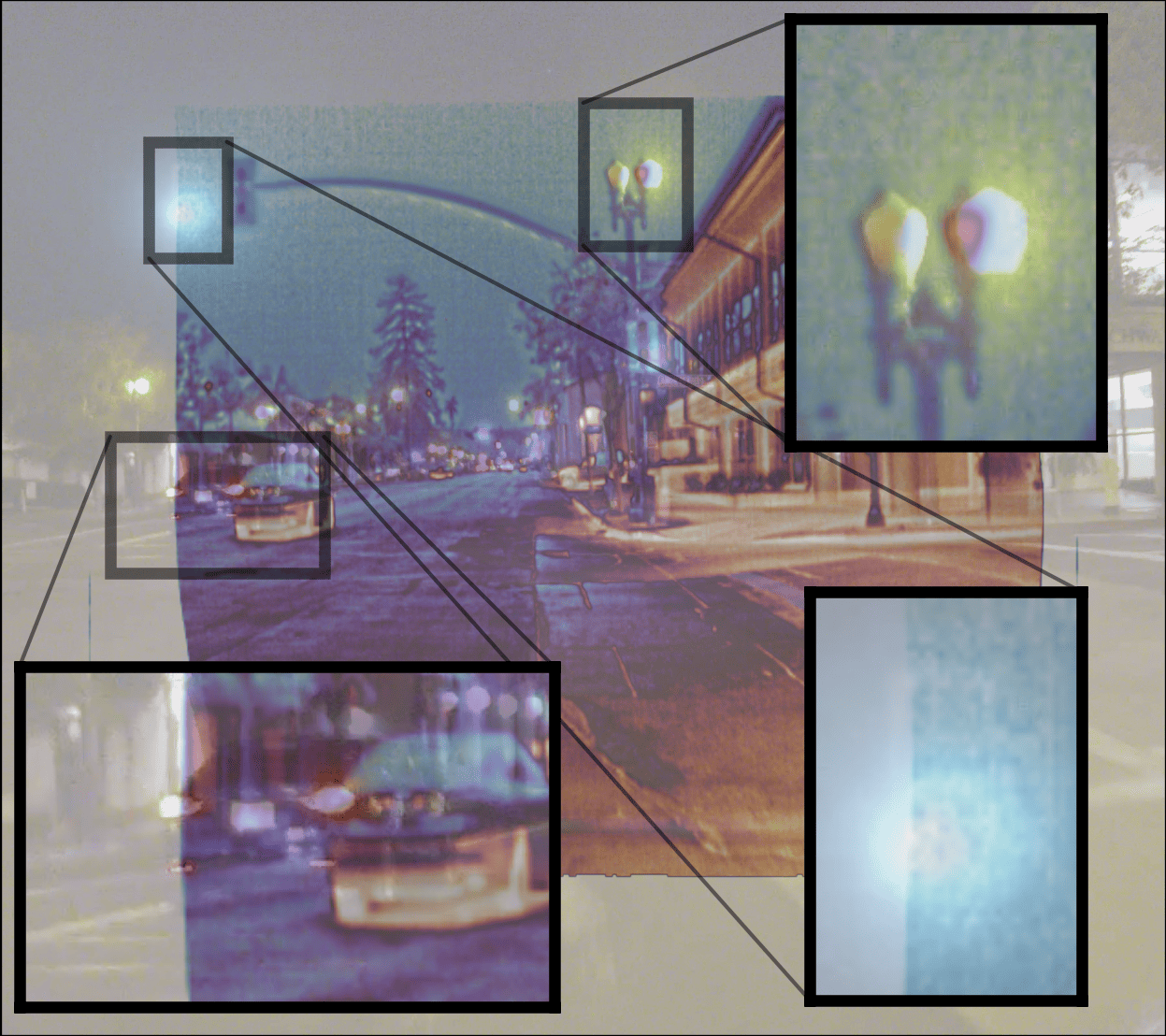} &
        \includegraphics[width=\tablength]{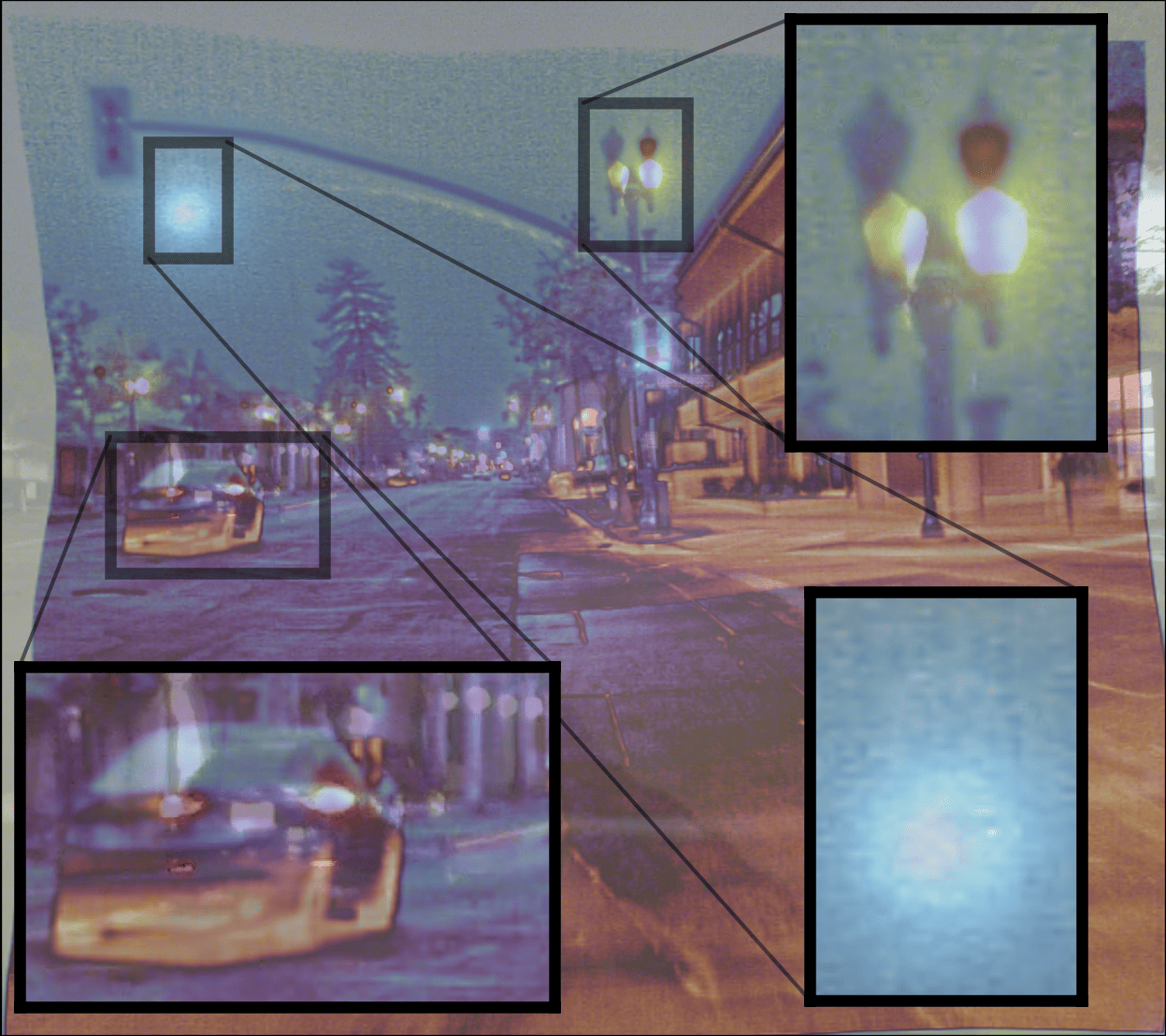} &
        \includegraphics[width=\tablength]{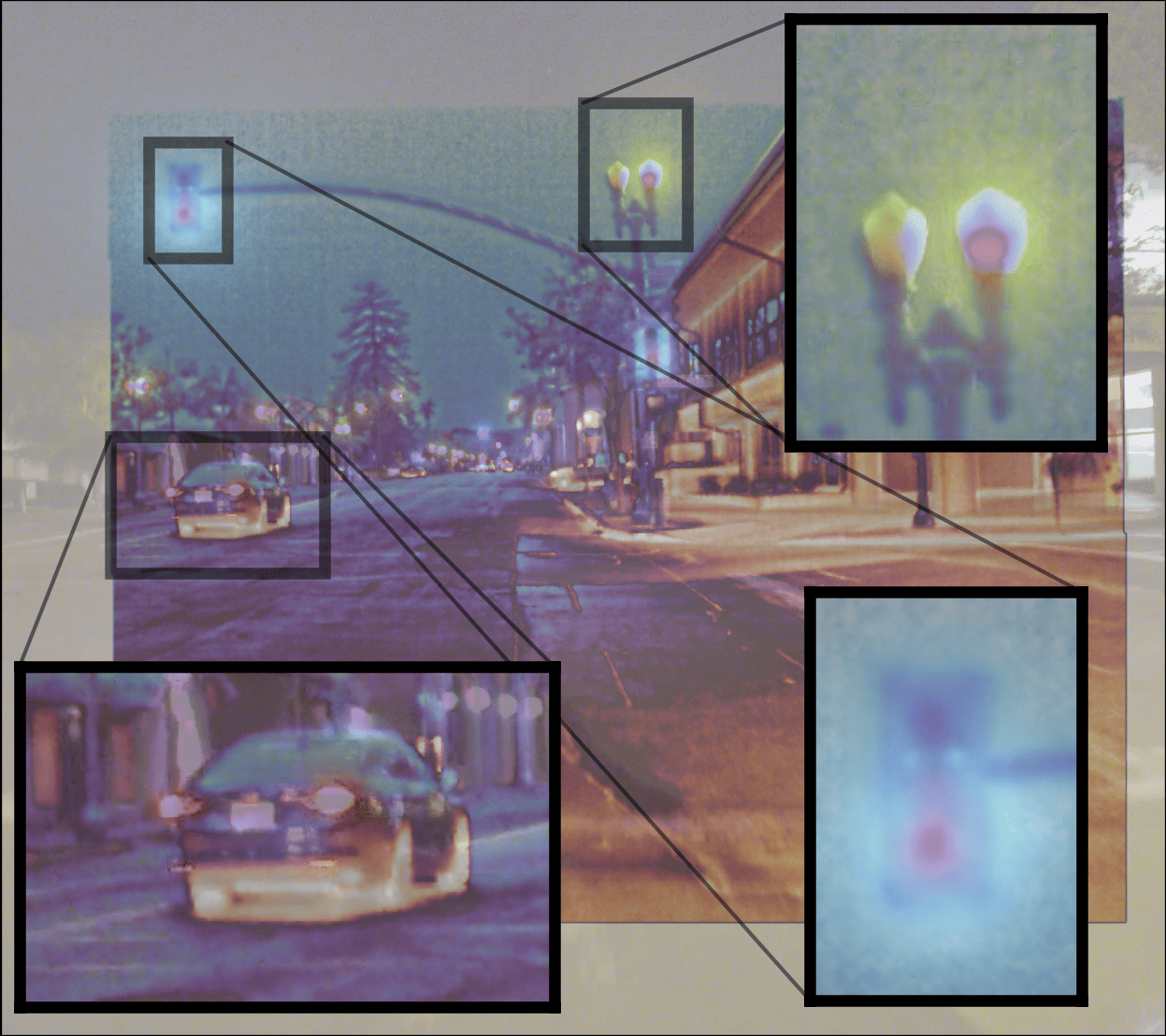}\\

    \end{tabularx}

    \caption{Superimposed images with details for each methods (columns) and each datasets (rows). 'Source' are the raw images superimposed without registration. In each fused image the infrared image is only colormapped with the colormap 'twilight' from Matplotlib. For the second row, the visible registered image and the visible ground truth are superimposed.}
    \label{fig:qualitative_results}
\end{figure*}

\subsubsection{Quantitative Results}

For each dataset, metrics are computed on image pairs cropped to their common \acrlong{fov}, excluding undefined pixels and ensuring a fair comparison. The \textit{Source} values are computed on the corresponding unregistered infrared and visible images after the same crop. The results are reported in Table~\ref{tab:quantitative_results}.

It is worth clarifying that the use of NEC both as an optimization objective and as an evaluation metric does not constitute a direct optimization of the final evaluation result. NEC provides gradients to optimize the camera parameters, which are constrained to represent a physically meaningful camera configuration. The registered image is subsequently obtained through the depth-dependent reprojection induced by these parameters. Thus, the optimization does not directly learn or manipulate an unconstrained alignment field to maximize NEC; rather, NEC guides the estimation of a compact set of camera parameters whose resulting reprojection is evaluated independently. This distinction is particularly important for interpreting NEC as a measure of the quality of the recovered geometric alignment.

The first key observation is that RIFT~\cite{reg_homo_RIFT} and our method provide the most consistent improvements across the considered datasets. RIFT achieves strong quantitative performance overall, but its global transformation cannot explicitly account for depth-dependent parallax. SuperFusion~\cite{reg_flow_superfus} and CrossRAFT~\cite{reg_flow_crossRaft} perform well on some datasets but are less robust to large geometric discrepancies between the sensors. In contrast, our method maintains strong performance across all datasets without post-processing the estimated reprojection.

\subsubsection{Qualitative Results}

Figure~\ref{fig:qualitative_results} presents representative examples from the original sample selection of each dataset, together with magnified views of selected regions.

Our method consistently produces well-aligned structures across the different datasets. In particular, the magnified regions illustrate its ability to recover pixel-level correspondences while accounting for depth-dependent parallax. This behavior is especially visible in regions containing objects at different depths, where a single global transformation is insufficient to achieve accurate alignment. Importantly, these results are obtained without post-processing the occlusion-induced artifacts, further demonstrating the robustness of the estimated camera geometry.

\subsection{Experiment 4. Multi-scale translation parametrization model}

\begin{figure}[htbp]
\centering
\includegraphics[width=\linewidth]{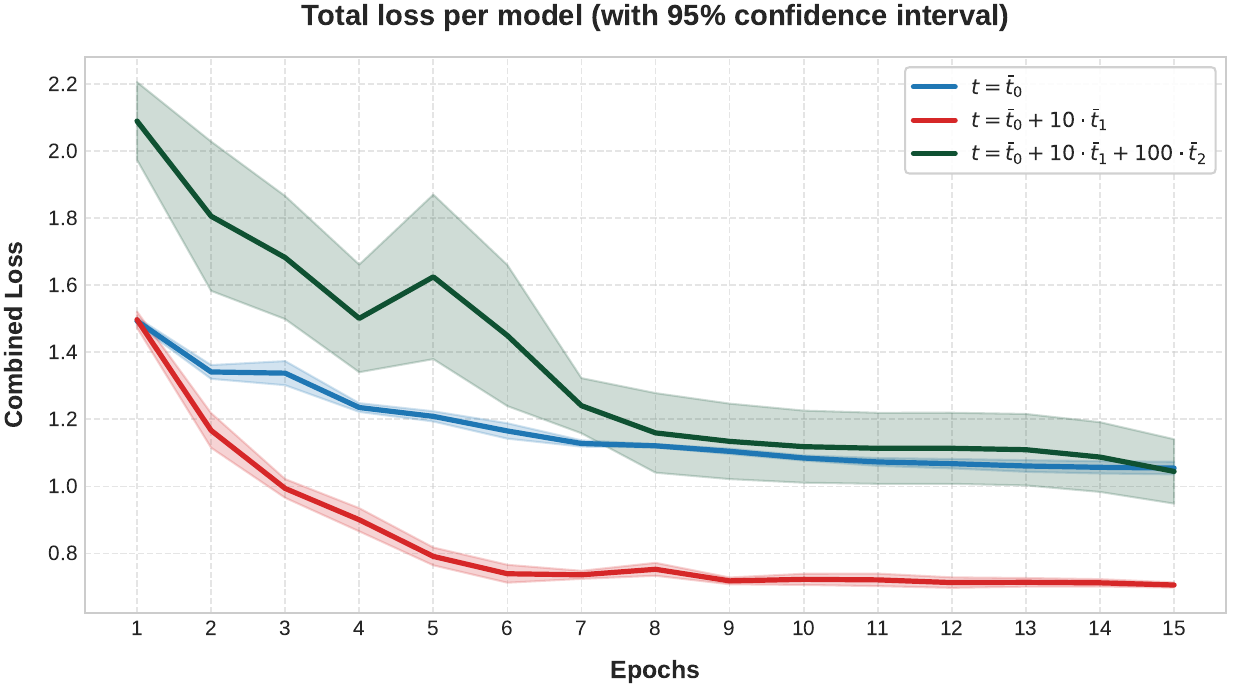}
\caption[Comparison of the convergence behavior of the optimization with different translation parameterizations]{Comparison of the convergence behavior of the optimization with different translation parameterizations. The combined loss is the weighted sum of the flow and image losses described in the following paragraph. The transparent shaded area around each curve represents the standard deviation across 10 randomly initialized runs, while the solid line represents the mean loss.}
\label{fig:translation_test}
\end{figure}

To assess the benefit of the proposed multi-scale translation parameterization, we compare optimization using different numbers of translation scales. Figure~\ref{fig:translation_test} reports the evolution of the combined loss over ten randomly initialized runs for each parameterization. The results show that introducing a coarse translation component consistently improves convergence compared with optimizing a single translation scale. In particular, the additional coarse degree of freedom facilitates the exploration of broader translation configurations, allowing the optimization to reach lower-loss solutions while retaining the fine-scale component for subsequent local refinement. In contrast, introducing a second coarse component does not provide a consistent improvement and instead leads to increased variability and less stable convergence across runs. This suggests that increasing the number of translation scales unnecessarily enlarges the optimization space and can hinder the gradient-based optimization. Therefore, the two-scale parameterization, combining one fine and one coarse translation component, provides the most favorable trade-off between global exploration and local refinement and is retained for the remainder of the experiments.

\section{Discussion and Conclusion}

\noindent In this work, we have presented XCalib, a multimodal image registration framework that constrains dense correspondences through a persistent camera model and scene depth. Rather than directly optimizing an unconstrained image-space deformation, the proposed approach derives the registration field from a compact set of camera parameters and monocular metric depth. This geometric formulation is particularly suited to scenes exhibiting depth-dependent parallax, where global transformations can be too restrictive while dense correspondence-based methods may produce less stable alignments. The experiments demonstrate that XCalib remains consistently competitive across diverse datasets, including daytime, nighttime, and large geometry-mismatch scenarios, while providing smoother frame-to-frame registration on image sequences.

Experimental evaluations show that the proposed method matches or surpasses state-of-the-art approaches across the considered datasets and metrics, while maintaining consistent qualitative alignment. A key strength is the persistence of the geometric model: once optimized for a camera rig, the same parameters can be reused across subsequent synchronized frames, providing an explicit constraint that promotes temporal consistency. The use of depth further allows the framework to account for non-uniform displacements caused by scene geometry without requiring a separate dense deformation to be estimated for every image pair. Moreover, the depth input has a limited influence on the estimation of the global registration parameters, making monocular metric depth a versatile source of geometric information even when its local predictions are imperfect.

The remaining influence of depth errors is primarily observed locally. Incorrect predictions around small structures or fine textures can introduce localized reprojection errors that may appear as slight texture instability or flickering in image sequences. A second limitation concerns computational cost, as monocular depth estimation represents the main computational bottleneck of the current pipeline and limits its direct use in real-time embedded applications. However, monocular depth estimation is not intrinsic to the proposed formulation. Sparse metric measurements from LiDAR or time-of-flight (ToF) sensors combined with learned depth completion or upscaling could provide an alternative, potentially enabling substantially faster dense depth estimation while retaining the geometric benefits of the framework.

\textbf{Future Work:}
Future research will focus on extending the framework toward real-time operation, in particular by investigating sparse depth sensing combined with efficient learned depth reconstruction, with the objective of reaching high frame rates suitable for embedded ADAS applications. Explicit treatment of occluded and disoccluded regions could further reduce local reprojection artifacts. Finally, although the optimized camera parameters are reasonably close to the physical configuration of the camera rig, the present work does not aim at precise camera calibration. The parameters primarily serve as a compact geometric representation from which the multimodal registration is derived. A more systematic comparison with accurately calibrated camera setups would nevertheless be valuable to further characterize the relationship between the optimized parameters and the underlying physical geometry.

\bibliographystyle{unsrt}  
\bibliography{library}  

\printglossaries

\end{document}